%% file: PetriBench.tex
\documentclass{article} 
\usepackage{iclr2027_conference,times}

\input{math_commands.tex}

\usepackage{hyperref}
\usepackage{url}
\usepackage{xcolor}
\usepackage{caption}
\usepackage{booktabs}
\usepackage{listings}
\usepackage[T1]{fontenc}
\usepackage[most]{tcolorbox}
\tcbuselibrary{listings,breakable,skins}
\usepackage{placeins}

\definecolor{pastelblue}{HTML}{A7C7E7}
\definecolor{pastelcoral}{HTML}{E6A4A4}
\definecolor{pastelgreen}{HTML}{A8D5BA}

\definecolor{softred}{RGB}{190,70,70}
\definecolor{softblue}{RGB}{65,105,170}

\usepackage{todonotes}

\title{PetriBench: Benchmarking LLM Reasoning over Dynamic State Spaces}

\author{%
\textbf{Pyrros Koussios\textsuperscript{1},
Benjamin Jäger\textsuperscript{1},
John Hua Yao\textsuperscript{2},
Ajay Sridhar\textsuperscript{2},} \\
\textbf{Violet Xiang\textsuperscript{2},
Chenhao Li\textsuperscript{1}} \\[6pt]
\textsuperscript{1}\,ETH Zürich \quad
\textsuperscript{2}\,Stanford University \\[3pt]
{\small\texttt{\{pkoussios,benjaeger,chenhli\}@ethz.ch}} \\
{\small\texttt{\{johnyao,ajaysri,ziyxiang\}@stanford.edu}}
}
\iclrfinalcopy 
\begin{document}

\maketitle

\begin{abstract}
Characterizing LLM reasoning remains an open challenge, as many existing benchmarks isolate specific reasoning skills, rely on external knowledge, or are costly to extend.
We introduce \emph{PetriBench}, a compact, fully self-contained, and scalable benchmark for evaluating LLM reasoning over dynamic state spaces using Petri nets, a mature formalism for modeling real-world concurrent and distributed systems.
PetriBench organizes reasoning into four task families varying by scope and temporal horizon, with Easy, Medium, and Hard levels generated by increasing structural complexity and evaluated against exact ground truth.
Across a diverse set of proprietary and open-weight models, accuracy decreases consistently with difficulty, while harder instances expose increasingly distinct task-specific capability profiles.
Additional analyses show that test-time compute improves performance but interacts differently with different reasoning tasks, and that procedural generation yields smooth scaling with structural complexity.
Together, these results show that PetriBench provides a unified and extensible setting for probing the strengths, limits, and scaling behavior of LLM reasoning.
\end{abstract}

\section{Introduction}\label{sec:introduction}
Large Language Models (LLMs) have advanced rapidly in recent years \citep{achiam2023gpt}.
As general-purpose systems, they are expected to operate across diverse domains, representations, and problem structures, making their capabilities difficult to characterize comprehensively \citep{liang2022holistic}.
While existing evaluations measure many important dimensions, such as safety \citep{zhang2024safetybench} and factual knowledge \citep{hendrycks2020measuring}, logical reasoning, as one of the most fundamental components of intelligence, remains challenging to be systematically formulated and assessed.

Most existing reasoning benchmarks express problems through natural language or manually curated examples \citep{cobbe2021trainingverifierssolvemath, suzgun2022challengingbigbenchtaskschainofthought}.
While such evaluations have driven considerable progress, extending them can be costly, and solving them often requires knowledge beyond what is explicitly specified in the prompt.
More abstract benchmarks reduce some of these concerns by evaluating models on formally defined structures.


Graph reasoning benchmarks, for example, evaluate connectivity, shortest paths, and other structural properties \citep{wang2024languagemodelssolvegraph}.
Other abstract environments introduce evolving state through finite-state machines, symbolic execution, or sequences of natural-language updates \citep{samiei2025illusionproceduralreasoningmeasuring, wu2025computationalreasoninglargelanguage,
rezaee2025exploringstatetrackingcapabilities}.
These settings capture important aspects of state-space reasoning, but typically emphasize a particular form of dynamics or reasoning objective.

Petri nets \citep{Petri1962} provide a unified formalism that brings many of these reasoning settings together.
Petri nets can be viewed as graphs with an explicit notion of state: vertices hold tokens, whose distribution over the net defines the current state of the system, while edges specify how this distribution may change, with each firing consuming or producing tokens at connected vertices and thereby updating the system state.
This compact representation supports reasoning over distributed state, alternative execution paths, shared resources, structural invariants, and behavior over unbounded horizons.
Importantly, Petri nets are a well-established formalism used to model real-world concurrent and distributed systems, with applications spanning software, communication protocols, hardware, manufacturing processes, and business workflows \citep{Petri1962, murata1989petri}.

Building on this formalism, we introduce \emph{PetriBench}, a procedurally generated benchmark for evaluating LLM reasoning over dynamic state spaces.
PetriBench organizes its reasoning demands along two axes, local versus global scope and finite versus infinite horizon, instantiated through six tasks.
A collection of each task is procedurally generated at Easy, Medium, and Hard difficulty levels by varying the amount of property-relevant and distractor structure, enabling systematic evaluation as reasoning demand increases.
Task instances are fully self-contained, with all information required for solving them specified in the system prompt, and remain compact despite inducing combinatorially large or even infinite state spaces.
With target properties admitting exact verification without human or LLM-based judging, we evaluate a broad set of proprietary and open-weight models, study how performance and reasoning efficiency vary across task types and structural complexity, and analyze the distinct capability profiles that emerge.

Overall, our contributions are threefold.
\textbf{(i)} First, we introduce \emph{PetriBench}, a compact, fully self-contained, and procedurally scalable benchmark built on Petri nets, a mature formalism for modeling real-world concurrent and distributed systems, to evaluate diverse forms of reasoning within a single framework.
\textbf{(ii)} Second, we formulate a controlled taxonomy of state-space reasoning spanning local and global scope and finite and infinite horizons, instantiated through six tasks and multiple difficulty levels to systematically probe reasoning across different task structures.
\textbf{(iii)} Third, we benchmark a broad set of proprietary and open-weight models and provide detailed analyses of capability profiles, reasoning efficiency, scaling behavior, and failure modes.

\section{Background}\label{sec:background}
\subsection{Petri Net Formalism}\label{ssec:pn_formalism}

\begin{figure}[t]
    \centering
    \includegraphics[width=1.0\textwidth]{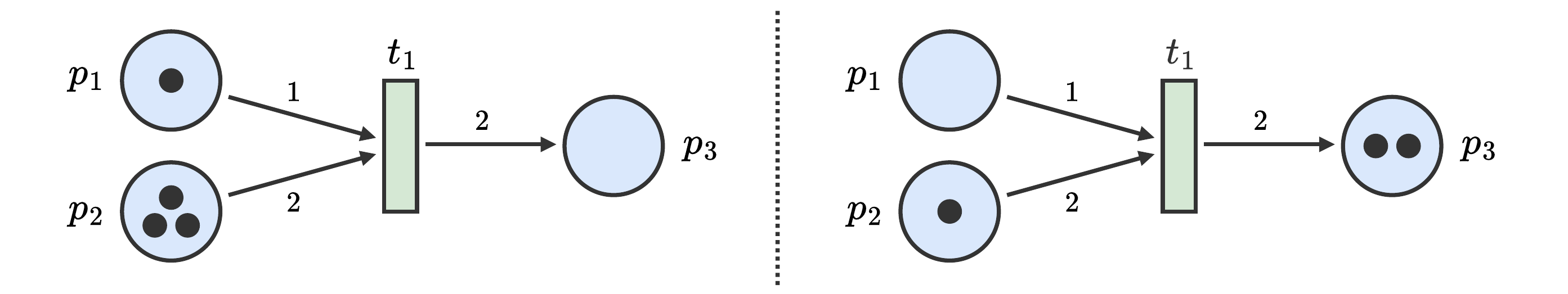}
    \caption{Illustration of a Petri net transition firing. The left and right panels show the marking before and after transition \(t_1\) fires, respectively. Places are shown in blue and tokens as black dots. Firing \(t_1\) consumes tokens from each input place \(p_1\) and \(p_2\) according to the corresponding input-arc weights and produces tokens in \(p_3\) according to the output-arc weight.}
    \captionsetup{justification=centering}
    \label{fig:pn_formalism}

    \vspace{-3.0ex}

\end{figure}

A Petri net is a directed bipartite graph describing how \emph{tokens} move between \emph{places} through \emph{transitions} \citep{Petri1962,murata1989petri}.
Formally, a weighted place-transition net is a tuple \[ \mathcal{N}=(P,T,F,W,M_0), \] where \(P=\{p_1,\ldots,p_m\}\) and \(T=\{t_1,\ldots,t_n\}\) are finite disjoint sets of places and transitions, \(F\subseteq(P\times T)\cup(T\times P)\) is the set of \emph{arcs}, and \(W:F\rightarrow\mathbb{N}_{>0}\) assigns their multiplicities.
We set \(W(x,y)=0\) whenever \((x,y)\notin F\).
A \emph{marking} \(M:P\rightarrow\mathbb{N}_0\) assigns a token count to every place and represents the state of the net, with \(M_0\) denoting the initial marking.

A transition \(t\) is enabled at \(M\) if \(M(p)\geq W(p,t)\) for every \(p\in P\).
Firing \(t\) produces the marking \[ M'(p)=M(p)-W(p,t)+W(t,p). \]
Executions proceed in discrete firing steps under standard interleaving semantics, with one enabled transition firing at each step.
A visual illustration of a Petri net transition firing is provided in Figure~\ref{fig:pn_formalism}.

Although Petri nets admit numerous extensions, including colored, timed, and stochastic variants, PetriBench confines itself to finite weighted place-transition nets with \(W(f)\in\{1,2\}\), where \(W(f)=1\) recovers the unweighted case.

\subsection{Petri Net Properties}\label{ssec:pn_properties}
A marking \(M\) is \emph{reachable} if some valid firing sequence transforms \(M_0\) into \(M\), written \(M_0\xrightarrow{\sigma}M\). A reachable marking is a \emph{deadlock} if it enables no transition. A net is \emph{bounded} if there exists \(k\in\mathbb{N}\) such that \(M(p)\leq k\) for every place \(p\) and every reachable marking \(M\).

A transition is \(L_0\)-live, or dead, if it cannot occur in any firing sequence from \(M_0\). It is \(L_4\)-live if, from every reachable marking, some continuation eventually enables it. These properties form the basis of the reasoning tasks introduced in Section~\ref{sec:petribench}.

\section{PetriBench}\label{sec:petribench}
\subsection{Task Taxonomy}\label{ssec:task_taxonomy}

\begin{figure}[t]
    \centering
    \includegraphics[width=1.0\textwidth]{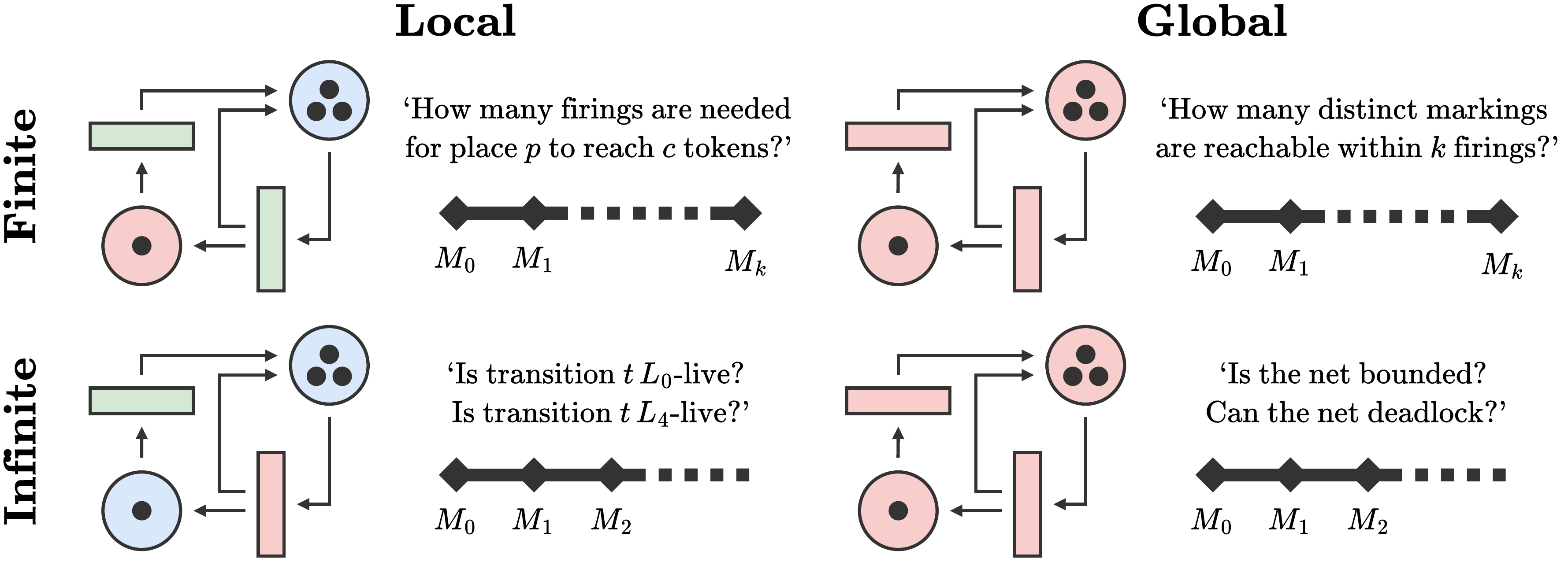}
    \caption{PetriBench organizes reasoning along two dimensions of scope and temporal extent.
    Together, they cover problems ranging from direct reasoning about finite state evolution to identifying invariants and long-term behavioral structure over the full system, all within a concise and well-studied formalism with broad real-world applicability.}
    \captionsetup{justification=centering}
    \label{fig:taxonomy}

    \vspace{-3.0ex}
\end{figure}

Petri nets admit a wide range of properties and associated analysis problems.
Consequently, an important design choice is how these properties are translated into reasoning tasks.
As different properties place different demands on reasoning, the selected tasks should ideally span a broad range of reasoning settings while remaining scalable and corresponding to questions that are practically relevant for systems modeled as Petri nets.

We propose a taxonomy of Petri net reasoning questions based on two dimensions, the scope of the queried property and its temporal extent.
The first axis distinguishes \emph{local} from \emph{global} questions.
A local question concerns a designated component of the net, such as a particular place or transition, whereas a global question concerns the behavior of the system as a whole, typically requiring reasoning over the complete reachable state space or all possible executions.
Importantly, local refers merely to the scope of the queried property, not to the information required to solve it.
Determining whether one transition is \(L_4\)-live, for example, may still require reasoning about the entire state space.

The second axis distinguishes \emph{finite-horizon} from \emph{infinite-horizon} questions.
Finite-horizon questions restrict reasoning to firing sequences of bounded length, where each firing transforms the current marking into a new marking.
They can in principle be answered by exploring a finite portion of the state space.
Infinite-horizon questions instead concern behavior over arbitrarily long firing sequences or across all markings reachable from the initial marking.
Such questions cannot generally be resolved by simulating a fixed number of firing steps and instead require identifying invariants, recurring behavior, terminal regions, or indefinitely productive cycles.

For each combination of local or global scope and finite or infinite temporal extent, we propose corresponding reasoning tasks, as visualized in Figure~\ref{fig:taxonomy}. PetriBench instantiates these four categories using questions with exact Boolean or integer answers.

\vspace{-1.0ex}\paragraph{Minimum Token Steps}
Given a target place $p$ and threshold $c$, the model must determine the minimum number of transition firings needed to reach a marking with $M(p)\geq c$.
This represents a local, finite-horizon task requiring target-directed search while accounting for token consumption, production, and optimality over alternative firing sequences.

\vspace{-1.0ex}\paragraph{Reachable Markings}
Given a firing limit $k$, the model must count the distinct markings reachable from $M_0$ within at most $k$ firings, including $M_0$.
This is a global, finite-horizon task requiring enumeration of branching executions while identifying when different firing sequences reach the same state.

\vspace{-1.0ex}\paragraph{Transition Liveness}
For a queried transition $t$, PetriBench asks both whether $t$ is $L_0$-live or $L_4$-live.
These are local, infinite-horizon properties: $L_0$ asks whether $t$ can ever fire, while $L_4$ requires that from every reachable marking some continuation can eventually fire $t$.
The two variants therefore distinguish permanent deadness from sustained liveness across all reachable states.

\vspace{-1.0ex}\paragraph{Deadlocks and Boundedness}
These tasks ask whether the net can reach a marking with no enabled transitions and whether token counts remain bounded over all reachable markings.
Both are global, infinite-horizon properties, testing whether any execution sequence terminates in a deadlock or can instead lead to arbitrarily large token counts in at least one place.

The proposed task set spans all four categories and covers a range of state-space reasoning problems.
We do not claim that this taxonomy captures every meaningful Petri net property or reasoning problem.
Rather, it provides a compact approximation that covers several practically relevant tasks while preserving deterministic ground truth and scalable evaluation.
Exact prompt templates and task-specific variations are provided in Appendix~\ref{app:prompts}.

\subsection{Property-Preserving Petri Net Generation}\label{ssec:pn_generation}

Each task is associated with a dedicated generation procedure that controls the structure determining the correct answer while allowing the surrounding Petri net to vary substantially.
Generated nets remain connected, with additional structure integrated through shared places, transitions, and cross-connections rather than isolated components.
Across generators, complexity is adjusted through structural expansion depth, connection density, initial token count, and task-specific parameters such as target firing count or execution horizon. Arc multiplicities are restricted to $W(f)\in\{1,2\}$.

\vspace{-1.0ex}\paragraph{Property-Preserving Construction}
For the local finite-horizon, local infinite-horizon, and global infinite-horizon tasks, the queried property is fixed by construction.
Local finite-horizon instances contain randomized dependency structures in which tokens must be accumulated through multiple paths or repeatable cycles before reaching a designated target.
Local infinite-horizon instances combine recurrent behavior with transitions that are permanently or eventually disabled.
Global infinite-horizon instances are constructed either around removable shared resources and indefinitely executable cycles or around structures that preserve a weighted token invariant or permit unbounded token growth.
Additional paths, cycles, transitions, and cross-connections are then introduced while preserving the queried property.
To guard against trivial generation artifacts, we verify that simple surface-level statistics do not reliably separate the target labels in Appendix~\ref{app:shortcut_baselines}.

The global finite-horizon task differs in that no predefined property must be preserved during construction, since the answer is determined by the induced state space.
We therefore start from a small cyclic Petri net with $\lvert P\rvert = 5$ and repeatedly apply randomized structural transformations that introduce longer sequences, alternative branches, cycles, and additional connections between existing components.
For a horizon of $k$ firings, we compute the exact number of reachable markings through bounded state-space exploration using the TINA solver \citep{berthomieu2004tool}, counting all distinct markings reachable within depth $k$.

\vspace{-1.0ex}\paragraph{Controlling Complexity}
Although each generator exposes task-specific parameters, two controls capture the principal sources of structural complexity.
\emph{Causal Generation Depth} scales the amount of property-determining structure and therefore the depth and complexity of the dependencies governing the correct answer.
The \emph{Distractor Generation Factor} controls the budget of additional label-preserving generation operations relative to this causal structure.
A factor of $0$ introduces no additional distractor generations, while a factor of $1$ assigns approximately equal generation budgets to causal and distractor structure.
Under our reachable markings construction, any additional fireable structure can alter the reachable-state set and is therefore answer-relevant by definition.
Restricting distractors to permanently disabled structure would sharply limit the diversity of admissible transformations while largely adding context without meaningful state-space interaction.
We therefore generate only fireable structure for this task, and do not define a separate distractor-generation axis.
Further generation details and parameters are provided in Appendix~\ref{app:generation}.

\subsection{Benchmark Composition}\label{ssec:benchmark_composition}

PetriBench comprises three difficulty tiers, \emph{easy}, \emph{medium}, and \emph{hard}, obtained by varying task-specific generation parameters while holding the task definitions and evaluation protocol fixed.
Each difficulty tier contains $1{,}600$ questions, evenly distributed across the four taxonomy categories defined in Section~\ref{ssec:task_taxonomy}.
The local finite-horizon and global finite-horizon subsets contain $400$ \texttt{MinimumTokenSteps} and $400$ \texttt{ReachableMarkings} questions, respectively.
The local infinite-horizon subset contains $400$ liveness questions, balanced across positive and negative $L_0$ and $L_4$ queries.
The global infinite-horizon subset contains $200$ \texttt{Deadlock} and $200$ \texttt{Boundedness} questions, with balanced positive and negative labels for each property.
Although all nets are designed to preserve the target properties by construction, every instance is independently verified using the TINA solver \citep{berthomieu2004tool} or counterexamples.

Petri nets are serialized using a compact edge-list representation allowing difficult state-space reasoning problems to be expressed compactly without confounding performance with long-context processing.
To remove construction-order artifacts, place and transition identifiers are randomly permuted before serialization.
All main results use chain-of-thought prompting with exact-match scoring against deterministic Boolean or integer targets.
All instances are generated from fixed random seeds, and we release the benchmark data, generators, and evaluation code.
We additionally ablate prompting without chain-of-thought and alternative serialization formats including PNML \citep{pnml} and JSON in Appendices~\ref{app:prompting} and \ref{app:serialization}.

\section{Results}\label{sec:results}
\subsection{Benchmark Performance}\label{ssec:benchmark_performance}

We evaluate a diverse set of proprietary and open-weight language models on PetriBench.
Table~\ref{tab:model_performance} reports exact-match accuracy at each difficulty level, averaged equally across the four reasoning categories.
Two of the four equally weighted taxonomy categories are balanced binary tasks, while the two exact-match integer categories have negligible chance accuracy, yielding an aggregate random-guess baseline of approximately \(25\%\).
Selected models are evaluated at their default reasoning effort, with additional reasoning-effort settings included for select state-of-the-art models to characterize test-time scaling behavior, as discussed further in Section~\ref{ssec:reasoning_efficiency}. Full inference configurations, including sampling parameters and output-token limits, are provided in Appendix~\ref{app:evaluation_protocol}.

Table~\ref{tab:model_performance} reveals a broad spectrum of capability across the evaluated models.
Across all models, performance decreases consistently as benchmark difficulty increases, while the separation between model families becomes increasingly pronounced.
Easy tasks are solved reliably by several frontier systems while still discriminating weaker models, whereas medium and hard tasks retain clear headroom even under stronger reasoning configurations.
Overall, PetriBench spans a broad range of difficulty across model capability levels rather than concentrating evaluation within a narrow performance regime.
Supplementary analyses of model behavior on PetriBench are provided in Appendices~\ref{app:petribench_model_diagnostics} and~\ref{app:error_analysis}.

\begin{figure*}[t]
    \centering
    \includegraphics[width=1.0\textwidth]{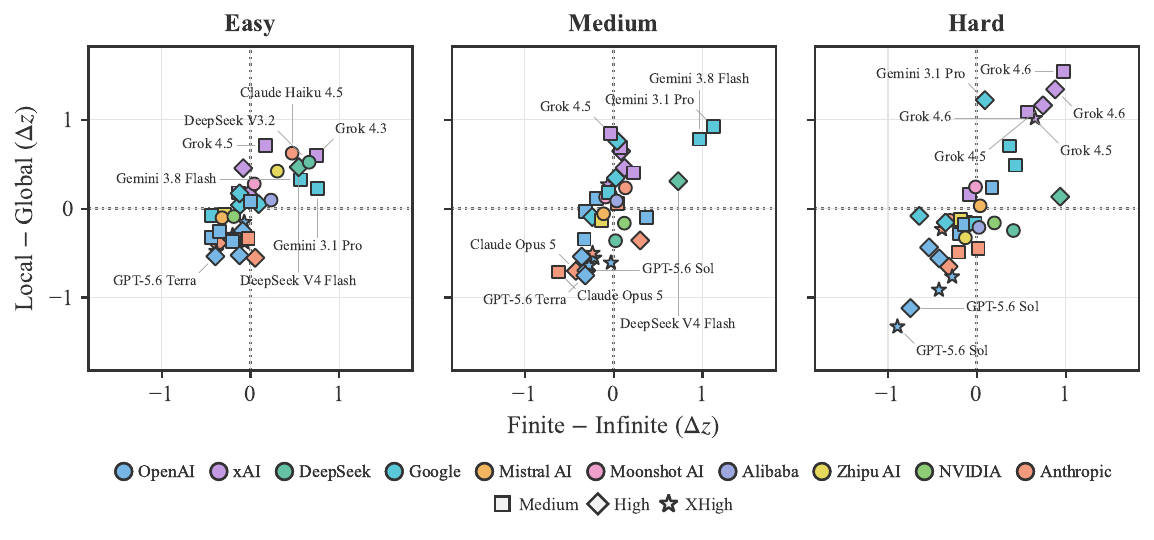}
    \caption{
    Task-specific capability profiles across PetriBench difficulty levels.
    Within each level, performance on the four taxonomy categories is standardized across evaluated model configurations and projected onto Finite--Infinite and Local--Global performance contrasts.
    Positive horizontal and vertical values indicate relative specialization toward finite-horizon and local reasoning, respectively.
    The increasing dispersion from Easy to Hard reflects increasingly pronounced task-specific differences in model capability.
    }
    \label{fig:capability_profiles}

    \vspace{-3.0ex}
\end{figure*}

Aggregate accuracy, however, becomes progressively less representative of \emph{task-specific} strength as benchmark difficulty increases.
Figure~\ref{fig:capability_profiles} standardizes performance relative to all other models, factors out differences in absolute task difficulty, and positions each model according to its finite--infinite and local--global contrasts in standardized accuracy.
Consequently, models in the upper-right quadrant exhibit comparatively stronger Local--Finite performance, while those in the lower-left exhibit comparatively stronger Global--Infinite performance, with the remaining quadrants interpreted analogously.

At lower difficulty, model profiles remain comparatively concentrated, indicating that performance is largely governed by a common notion of overall model strength.
As difficulty increases, the profiles spread out substantially, revealing increasingly pronounced task-specific strengths and weaknesses.
This trend is also quantified by the variance decomposition in Table~\ref{tab:difficulty_decomposition}.
As task difficulty increases, the first principal component, representing a shared axis of overall model performance, explains a progressively smaller fraction of performance across the individual tasks.
Thus, harder PetriBench instances increasingly differentiate models in ways that are not captured by overall performance alone, with task-specific variation becoming more pronounced as structural difficulty increases despite a strong shared component of model capability.

Lastly, we validate PetriBench against the Artificial Analysis Intelligence Index \citep{artificialanalysis}.
Aggregate PetriBench rankings achieve a Spearman correlation of $\rho=0.94$ across $n=28$ shared model configurations, providing strong external validation that PetriBench captures broadly recognized differences in model reasoning capability rather than a benchmark-specific ordering.
Further analysis is provided in Appendix~\ref{app:benchmark_correlation}.

\begin{table*}[t]
    \caption{Exact-match accuracy (\%) on PetriBench across difficulty levels.
    Reasoning effort is indicated by M (Medium), H (High), and XH (XHigh). Best proprietary and open-weight results are highlighted in red and blue, respectively.
    Random guessing yields a baseline of $25\%$.}
    \label{tab:model_performance}
    \centering
    \small
    \setlength{\tabcolsep}{3pt}
    \begin{tabular}{@{}lccc@{\hspace{1.5em}}lccc@{}}
        \toprule
        Model & Easy & Medium & Hard &
        Model & Easy & Medium & Hard \\
        \midrule
        Claude Opus 5 (XH)   & $92.7$ & $83.8$ & $71.7$ &
        GPT-5.6 Luna (XH)     & $68.9$ & $50.1$ & $34.4$ \\
        Claude Opus 5 (H)    & $90.4$ & $81.9$ & $71.8$ &
        GPT-5.6 Luna (H)      & $53.9$ & $36.4$ & $28.0$ \\
        Claude Opus 5 (M)    & $84.3$ & $70.0$ & $58.1$ &
        GPT-5.6 Luna (M)      & $38.8$ & $28.3$ & $24.6$ \\
        Grok 4.6 (XH)        & $95.3$ & $80.9$ & $56.5$ &
        Grok 4.3 (M)          & $50.0$ & $33.6$ & $26.6$ \\
        Grok 4.6 (H)         & $95.1$ & $76.5$ & $55.0$ &
        Claude Sonnet 5 (H)   & $57.4$ & $33.5$ & $27.1$ \\
        Grok 4.6 (M)         & $93.9$ & $75.7$ & $52.6$ &
        Claude Sonnet 5 (M)   & $49.6$ & $30.6$ & $25.5$ \\
        GPT-5.6 Sol (XH)     & $96.4$ & \textcolor{softred}{$\bm{90.9}$} & \textcolor{softred}{$\bm{77.8}$} &
        Claude Haiku 4.5      & $46.9$ & $31.1$ & $25.5$ \\
        GPT-5.6 Sol (H)      & $92.9$ & $81.4$ & $59.8$ &
        DeepSeek V4 Flash (H) & \textcolor{softblue}{$\bm{57.8}$} & \textcolor{softblue}{$\bm{34.0}$} & \textcolor{softblue}{$\bm{27.3}$} \\
        GPT-5.6 Sol (M)      & $82.7$ & $52.5$ & $31.3$ &
        Gemma 4 31B (H)       & $38.4$ & $26.1$ & $25.8$ \\
        Gemini 3.8 Flash (H) & \textcolor{softred}{$\bm{96.8}$} & $74.5$ & $41.3$ &
        Gemma 4 31B (M)       & $27.9$ & $24.4$ & $23.8$ \\
        Gemini 3.8 Flash (M) & $86.6$ & $58.7$ & $38.5$ &
        GLM-5.3 Flash (M)     & $33.5$ & $26.0$ & $24.9$ \\
        Grok 4.5 (H)         & $88.2$ & $67.2$ & $50.1$ &
        GLM-5                  & $34.6$ & $26.3$ & $24.3$ \\
        Grok 4.5 (M)         & $83.8$ & $65.0$ & $48.9$ &
        DeepSeek V3.2          & $39.1$ & $23.5$ & $21.3$ \\
        GPT-5.6 Terra (XH)   & $85.0$ & $64.4$ & $50.7$ &
        GPT-OSS-120B (M)      & $32.4$ & $25.6$ & $24.4$ \\
        GPT-5.6 Terra (H)    & $67.5$ & $44.3$ & $31.5$ &
        Kimi K2.5              & $32.8$ & $25.1$ & $24.4$ \\
        GPT-5.6 Terra (M)    & $53.9$ & $30.7$ & $25.3$ &
        Qwen3-235B-A22B        & $28.3$ & $23.7$ & $23.5$ \\
        Gemini 3.1 Pro (H)   & $92.3$ & $69.9$ & $46.9$ &
        Mistral Large 3        & $26.7$ & $24.5$ & $23.8$ \\
        Gemini 3.1 Pro (M)   & $75.5$ & $48.8$ & $34.8$ &
        Nemotron 3 Super       & $23.6$ & $22.8$ & $22.6$ \\
        \bottomrule
    \end{tabular}
\end{table*}

\begin{table}[t]
    \caption{
Variance explained by the first principal component of task-level model performance across difficulty levels.
Comparison columns report paired-bootstrap changes in explained variance with 95\% confidence intervals.
The decreasing explained variance statistically supports the increasing dispersion of model capability profiles observed in Figure~\ref{fig:capability_profiles}.
}
    \label{tab:difficulty_decomposition}
    \centering
    \small
    \setlength{\tabcolsep}{5pt}
    \begin{tabular}{ccccc}
        \toprule
        Easy & Easy $\rightarrow$ Medium & Medium & Medium $\rightarrow$ Hard & Hard \\
        \midrule
      $\mathbf{91.7}\text{\bfseries\%}$
      & $-4.4$ pp $[-6.4,-3.2]$
      & $\mathbf{87.3}\text{\bfseries\%}$
      & $-8.4$ pp $[-12.2,-6.4]$
      & $\mathbf{78.9}\text{\bfseries\%}$ \\
      \bottomrule
    \end{tabular}

    \vspace{-2.0ex}
\end{table}


\subsection{Reasoning Efficiency}\label{ssec:reasoning_efficiency}

\begin{figure}[t]
    \centering
    \includegraphics[width=1.0\textwidth]{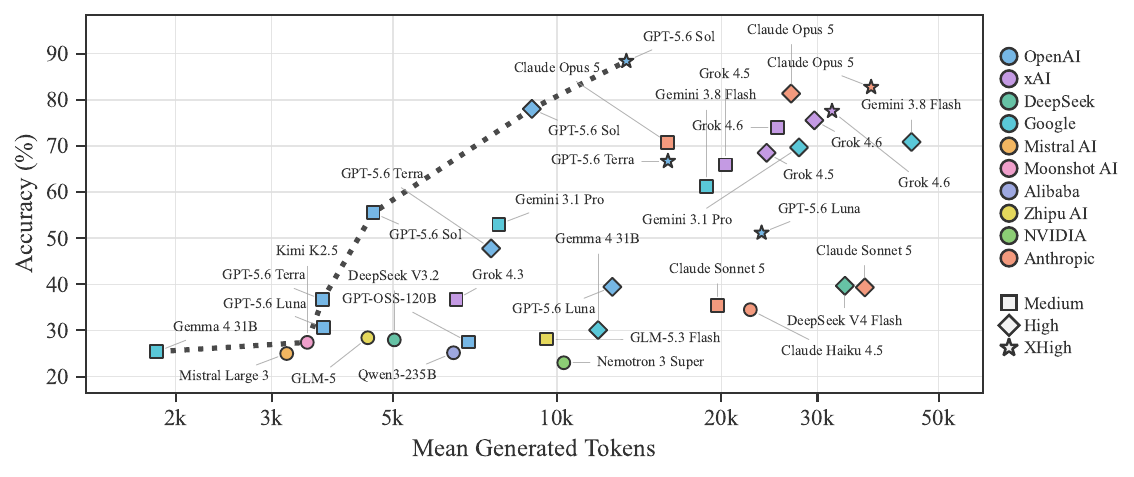}
    \caption{Accuracy versus reasoning-token Pareto frontier on PetriBench across evaluated model configurations.
    Aggregate accuracy is computed by equally averaging across task categories and difficulty levels, while the horizontal axis reports mean generated reasoning tokens.
    The resulting frontier spans a broad spectrum of performance and reasoning-token expenditure across models and effort settings.
}
    \label{fig:pareto}

    \vspace{-1.0ex}
\end{figure}

Model comparisons on reasoning tasks are inherently coupled to test-time compute, as differences in inference budget can induce substantial changes in accuracy \citep{snell2024scaling, muennighoff2025s1}.
Figure~\ref{fig:pareto} presents the accuracy versus compute Pareto frontier for PetriBench, relating aggregate accuracy to the mean number of generated reasoning tokens over all difficulty levels.
Model configurations span a wide range of test-time compute, with similarly labeled reasoning-effort settings often corresponding to substantially different token budgets across model families.
Moreover, the accuracy gains obtained from additional reasoning budget vary markedly across models, reflecting substantial differences in reasoning efficiency.
Overall, however, increasing reasoning effort consistently improves performance, albeit with varying efficiency and diminishing returns at higher compute.

\begin{figure}[t]
    \centering
    \includegraphics[width=1.0\textwidth]{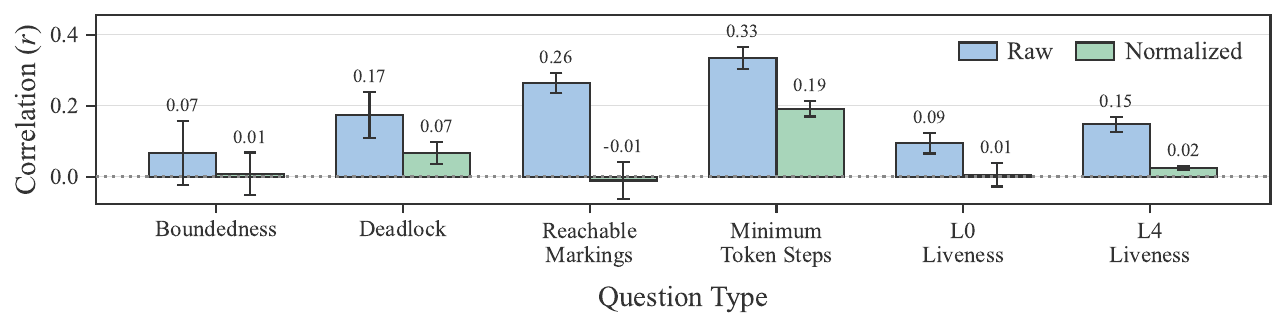}
    \caption{
    Correlation between reasoning length and binary correctness across PetriBench tasks, reported as mean Pearson correlation across difficulty levels with one standard deviation.
    The \emph{Raw} correlations control for benchmark difficulty, while the \emph{Normalized} correlations additionally remove systematic variation across models and task instances.
    The results separate finite- from infinite-horizon tasks and, after normalization, further distinguish the reasoning behavior of \texttt{MinimumTokenSteps} and \texttt{ReachableMarkings}.
}
    \label{fig:reasoning_correlation}

    \vspace{-2.0ex}
\end{figure}

The aggregate relationship between computation and performance, however, does not hold uniformly at the individual-question level.
Figure~\ref{fig:reasoning_correlation} reports the difficulty-controlled correlation between reasoning length and accuracy for each PetriBench task.
Longer reasoning is positively associated with correctness across all question types, with the strongest relationships observed for the two finite-horizon tasks.
This is consistent with the explicit search and state tracking required by \texttt{ReachableMarkings} and \texttt{MinimumTokenSteps}, in contrast to the more structural pattern recognition and invariant detection involved in the infinite-horizon property tasks, yielding a clear difference in reasoning-length behavior between finite- and infinite-horizon tasks.

Interestingly, Figure~\ref{fig:reasoning_correlation} shows that after additionally normalizing over the task set, the association largely vanishes for most question types.
In other words, for a given model, longer responses are generally no more likely to be correct than shorter ones.
Although both finite-horizon tasks exhibit a strong aggregate association between reasoning length and accuracy, only \texttt{MinimumTokenSteps} retains a pronounced positive correlation at the individual-response level.
This further differentiates the two finite-horizon tasks: additional reasoning length is not systematically associated with success for the breadth-oriented enumeration required by \texttt{ReachableMarkings}, whereas it remains positively associated with successful depth-oriented search toward a designated target in \texttt{MinimumTokenSteps}.
Together, these results suggest that additional test-time computation is broadly beneficial at the model level, while how that computation relates to successful reasoning depends strongly on the structure of the task.

\subsection{Difficulty Scaling}\label{ssec:difficulty_scaling}

\begin{figure*}[t]
    \centering
    \includegraphics[width=1.0\textwidth]{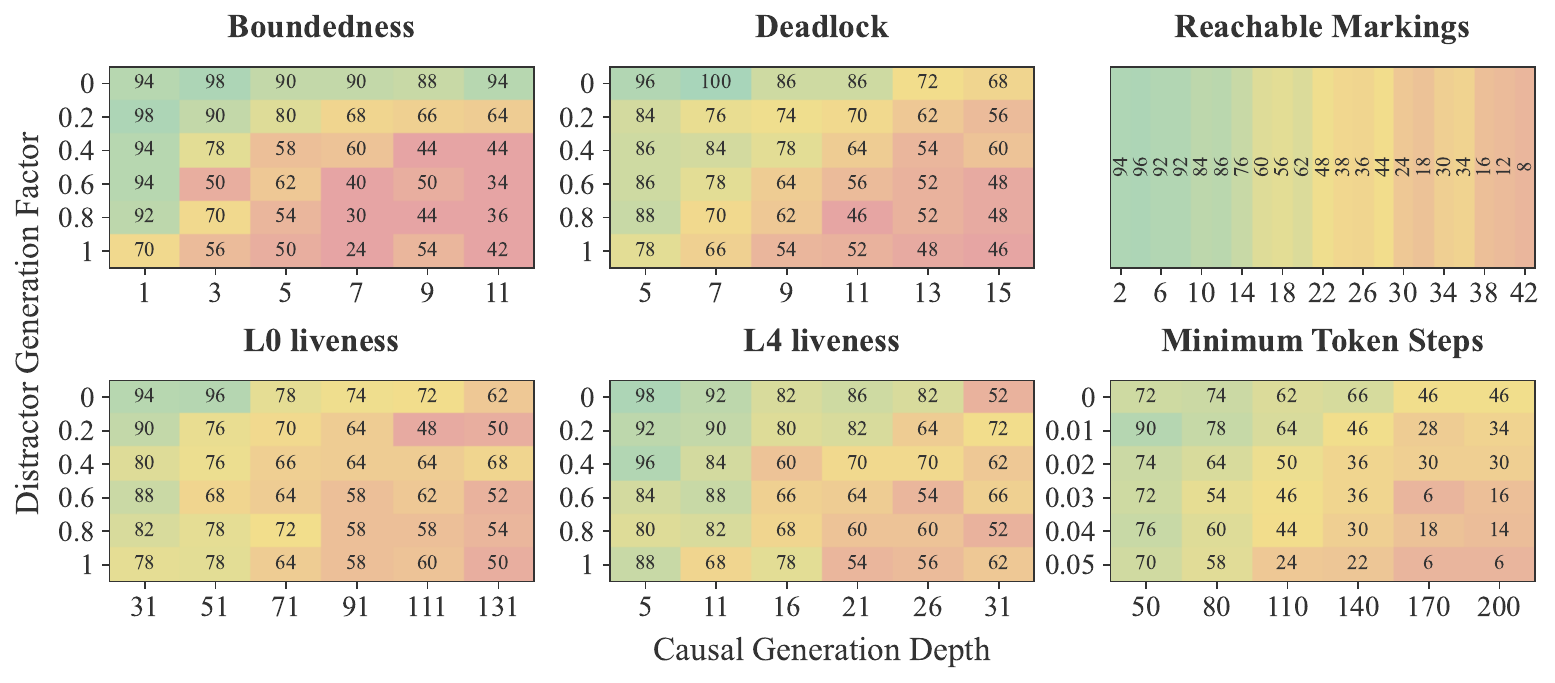}
    \caption{Difficulty scaling of GPT-5.6 Sol across PetriBench tasks, with $n=50$ instances per cell.
    Accuracy is reported as a function of \emph{Causal Generation Depth} and \emph{Distractor Generation Factor}, revealing systematic degradation with increasing complexity and distinct sensitivities across tasks.
    Binary tasks have a 50\% chance baseline, while integer-valued tasks are evaluated by exact match.}
    \label{fig:difficulty_scaling}

    \vspace{-3.0ex}
\end{figure*}

Figure~\ref{fig:difficulty_scaling} shows GPT-5.6 Sol (Medium) performance across the causal and distractor generation controls for each PetriBench task, with $n=50$ instances per cell, revealing substantial and mostly smooth performance degradation as structural complexity increases.
The task-specific scaling ranges further illustrate the different structural regimes required to challenge the same model, with properties such as boundedness becoming difficult after relatively limited property-determining generation, while \texttt{MinimumTokenSteps} remains tractable over substantially deeper generated structures.
As individual generation operations are task-specific, we interpret these results as within-task scaling trends rather than directly comparing sensitivity across question types.
Together, controllable structural complexity and efficient procedural generation highlight why Petri nets are particularly well suited for controlled evaluation of model reasoning capabilities.
We additionally verify these scaling trends using Gemini 3.8 Flash and demonstrate the richer, model-specific effects of causal and distractor complexity in Appendix~\ref{app:difficulty_scaling}.

\section{Related Work}\label{sec:related_work}

\vspace{-1.0ex}\paragraph{Petri Net Reasoning with Large Language Models}
Recent work has explored the use of LLMs for process mining and Petri net analysis.
PM-LLM-Benchmark \citep{berti2024pmllmbenchmarkevaluatinglargelanguage} evaluates models across a broad collection of process-mining tasks, including questions involving Petri nets, but primarily targets process-mining knowledge and model comprehension, with many open-ended responses evaluated using LLM judges rather than deterministic ground truth.
More recently, \citet{hu2026twostagereflectionrepromptingframework} study LLM-based Petri net reachability in an industrial setting, evaluating whether models can construct feasible firing sequences across six fixed Petri net structures.
Whereas these works study LLMs in specific Petri net applications, PetriBench uses the formalism itself as a medium for evaluating general reasoning over distributed state and state-space properties.

\vspace{-1.0ex}\paragraph{Reasoning Benchmarks and Formal Structures}
Many prominent evaluations of LLM reasoning focus on mathematical problem solving, scientific reasoning, or broad collections of general reasoning tasks \citep{cobbe2021trainingverifierssolvemath, hendrycks2021measuring, suzgun2022challengingbigbenchtaskschainofthought, rein2023gpqa, chollet2025arc}.
These benchmarks have been central to measuring advances in reasoning, but rely on prior domain knowledge, manually curated problem sets, or task distributions that are difficult to scale systematically.
Complementary work therefore evaluates reasoning in more explicitly defined computational environments.
Graph-based benchmarks such as NLGraph \citep{wang2024languagemodelssolvegraph}, GraphArena \citep{tang2025grapharena}, and GraCoRe \citep{yuan2025gracore} study structural and algorithmic reasoning over graph representations, including connectivity, shortest paths, flow, and combinatorial optimization.
Beyond graphs, related work evaluates finite-horizon execution and state tracking over Turing machines \citep{wu2025computationalreasoninglargelanguage}, finite-state machines \citep{samiei2025illusionproceduralreasoningmeasuring}, algebraic state transformations \citep{kim2023entity}, multi-entity state updates expressed in natural language \citep{rezaee2025exploringstatetrackingcapabilities}, and evolving symbolic game states such as chess \citep{kolasani2025llmchessbenchmarkingreasoning}.
A smaller body of work considers properties that require reasoning beyond a fixed execution horizon.
Program-verification approaches study the synthesis of inductive invariants that summarize behavior across arbitrary loop iterations \citep{kamath2024leveraging, wei2025invbench}, while program-termination benchmarks directly evaluate whether models can determine global properties of unbounded execution \citep{sultan2026halting}.
PetriBench brings these previously separate reasoning settings under a common compact, fully self-contained, and procedurally scalable framework, while grounding the evaluation in a formalism with broad applicability to real-world systems.

\section{Conclusion}\label{sec:conclusion}

We introduce \emph{PetriBench}, a compact, fully self-contained, and scalable benchmark for evaluating LLM reasoning over dynamic state spaces through Petri nets \citep{Petri1962}, a mature formalism for modeling real-world concurrent and distributed systems.
Through six procedurally generated tasks spanning local and global scope as well as finite and infinite horizons, PetriBench provides a unified evaluation across multiple reasoning demands with deterministic ground truth and controllable structural complexity.
We release all benchmark data, evaluation code, and model outputs, and maintain an online leaderboard for continued evaluation as new models become available.

Across a broad set of proprietary and open-weight models, performance degrades consistently with increasing difficulty while retaining substantial separation between model capabilities.
At the same time, harder instances reveal increasingly distinct task-specific capability profiles, suggesting that aggregate accuracy alone obscures meaningful differences in how models handle different forms of state-space reasoning.
Our analyses further show that test-time compute is an important but incomplete explanation of reasoning performance.
Greater reasoning effort generally improves accuracy, yet the relationship between reasoning length and correctness varies substantially across tasks.
Controlled generation experiments additionally demonstrate smooth scaling with structural complexity, providing a direct mechanism for increasing the difficulty as model capabilities improve.
Future work can extend PetriBench with additional reasoning questions and a broader distribution of procedurally generated net structures while maintaining target properties.

Taken together, these results highlight Petri nets as a powerful basis for controlled reasoning evaluation and position PetriBench as a framework for characterizing model reasoning capabilities across diverse structures, difficulties, and inference budgets.

\newpage

\subsubsection*{AI Use Statement}
Generative AI tools (OpenAI Codex) were used to assist with implementation, including code for Petri net generation and to improve consistency in result visualization and plotting.
Generative AI was also used to assist with polishing and editing of the manuscript.
The research questions, benchmark design, experimental protocol, execution of experiments, and interpretation of results were carried out by the authors.
We additionally use Codex in the qualitative error analysis in Appendix~\ref{app:error_analysis} to assist in constructing the failure-mode taxonomy and classifying sampled errors, with the goal of reducing human bias that can arise when categories are formed from only the subset of failures that can be inspected and retained in human memory at once.
The procedure and associated assumptions are described in detail in that section.
All AI-assisted code has been reviewed by the authors and visualizations were checked against the underlying experimental results.
We take responsibility for the final content of this work, including all text, claims, code, and artifacts produced with the aid of generative AI.

\subsubsection*{Ethics Statement}
This work does not involve human subjects, personal information, or sensitive data.
PetriBench is constructed from procedurally generated Petri nets and evaluated using automatically verifiable ground-truth answers, avoiding the need for human annotation or subjective judging.
The benchmark is intended for evaluating and analyzing language-model reasoning capabilities and does not introduce capabilities for autonomous action or deployment in safety-critical settings.
Potential risks include unintended use of the openly released benchmark instances, generators, or evaluation artifacts as training data, which could contaminate future evaluations and inflate reported performance.
We therefore encourage contamination-aware evaluation using newly generated held-out instances.
The benchmark otherwise presents limited direct ethical risk.

\subsubsection*{Reproducibility Statement}
We describe all benchmark generation, evaluation, and scoring procedures in detail in Appendix~\ref{app:implementation_details} to support full reproduction of our results.
Additionally, we release all materials required to reproduce PetriBench and the analyses in this work, including the full benchmark, complete generation and evaluation code, and reproducible procedures for generating new instances under the same task definitions and difficulty controls.
We also release all model outputs from the reported experiments, including reasoning traces, parsed predictions, and failure-mode annotations.
Finally, we maintain a public website with updated PetriBench results and encourage external evaluation and community contributions to the benchmark.

\bibliography{iclr2027_conference}
\bibliographystyle{iclr2027_conference}

\appendix
\section{Additional Results}
\subsection{Petri Net Serialization}\label{app:serialization}
Petri nets are defined as a mathematical formalism rather than by a canonical textual representation, making serialization an important design choice that may directly affect LLM reasoning performance.
We therefore evaluate two proprietary and two open-weight models on a fixed randomly sampled quarter of PetriBench, using the same instances across the compact edge-list representation, JSON, and PNML \citep{pnml}, an XML-based standardized interchange format for Petri nets.

\begin{figure*}[t]
    \centering
    \includegraphics[width=1.0\textwidth]{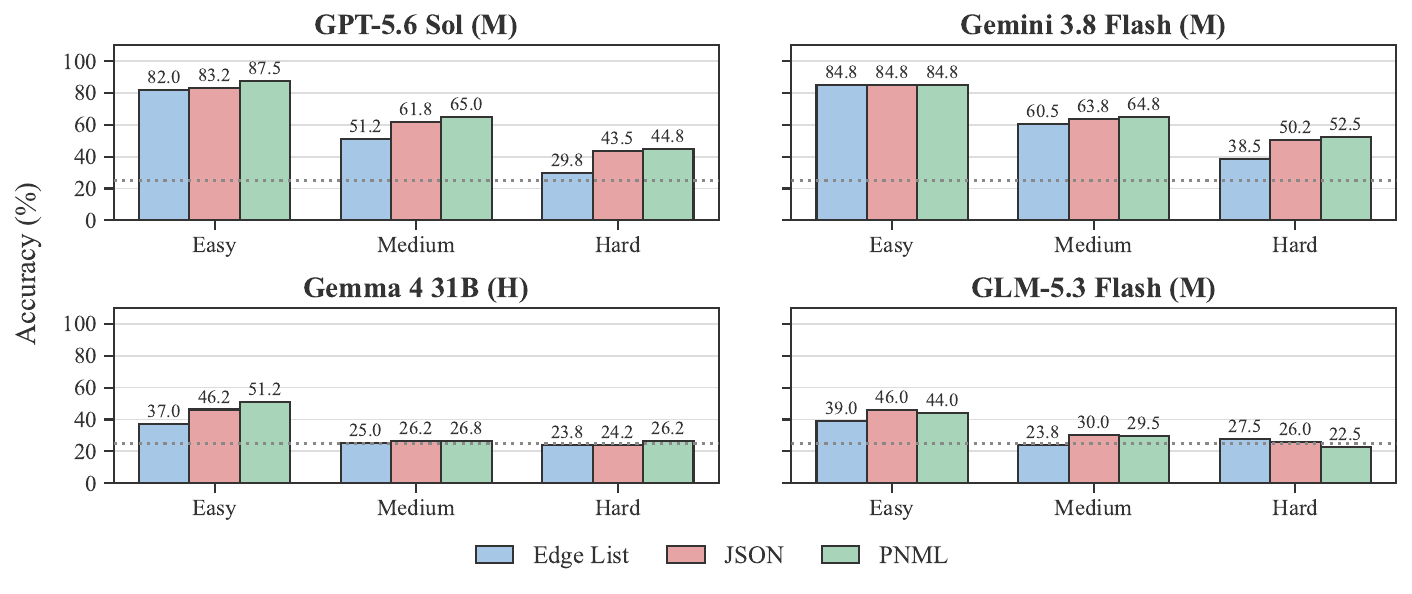}
    \caption{Exact-match accuracy under compact edge-list, JSON, and PNML \citep{pnml} serialization for two proprietary and two open-weight models on a fixed randomly sampled quarter of PetriBench.
    Results are reported across difficulty levels and show that serialization slightly affects absolute performance while consistently preserving the benchmark's difficulty progression across model families.}
    \label{fig:serialization}
\end{figure*}

Figure~\ref{fig:serialization} shows that, although serialization affects absolute performance, the underlying difficulty progression remains highly consistent across representations and model families.
Averaged across the four models, accuracy decreases from easy to medium to hard under the edge-list representation ($60.7\%\rightarrow40.1\%\rightarrow29.9\%$), JSON ($65.1\%\rightarrow45.5\%\rightarrow36.0\%$), and PNML ($66.9\%\rightarrow46.5\%\rightarrow36.5\%$).
The same ordering is largely preserved across serialization formats, becoming inconsistent only as model performance approaches the $25\%$ random-guessing baseline.
Thus, while more explicit representations generally improve absolute accuracy, the average gain remains modest at no more than $6.4\%$, and the benchmark’s relative difficulty structure remains stable across substantially different textual encodings.

\begin{figure*}[t]
    \centering
    \includegraphics[width=0.8\textwidth]{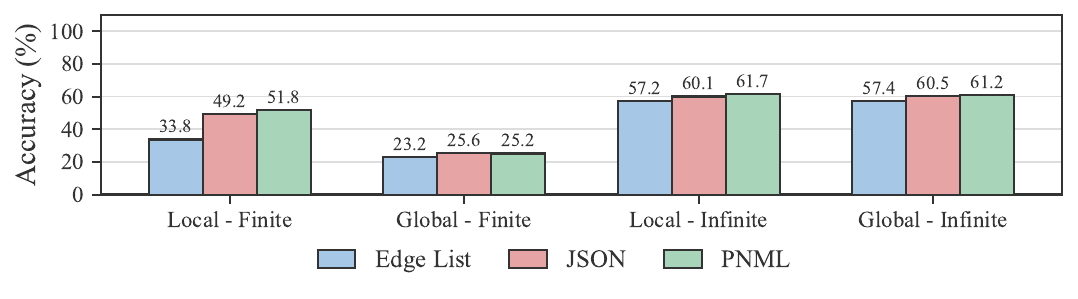}
    \caption{Exact-match accuracy under compact edge-list, JSON, and PNML \citep{pnml} serialization, averaged across evaluated models and difficulty levels.
    Serialization has a comparatively small effect on most task families, while the local finite-horizon \texttt{MinimumTokenSteps} task exhibits substantially greater sensitivity to representation.}
    \label{fig:serialization_averaged}
\end{figure*}

Serialization nevertheless affects absolute performance in a task-dependent manner.
Figure~\ref{fig:serialization_averaged}, which aggregates results across models and difficulty levels, shows that the local finite-horizon \texttt{MinimumTokenSteps} task represents the outlier.
This behavior is consistent with the distinct search structure of \texttt{MinimumTokenSteps}.
While \texttt{ReachableMarkings} emphasizes breadth-oriented exploration over a short horizon, \texttt{MinimumTokenSteps} requires depth-oriented reasoning toward a designated target, for which the more explicit structure of JSON and PNML appears particularly beneficial.
Overall, these results align with prior work showing that serialization effects can be highly task-dependent \citep{fatemi2023talklikegraphencoding,herbst2026lostserializationinvariancegeneralization}.

Although JSON and PNML generally improve absolute accuracy, these gains come at the cost of substantially longer prompts, as shown in Table~\ref{tab:serialization_length}.
We therefore retain the compact edge-list as the default representation for the main evaluation, as it provides the most compact encoding while preserving the benchmark's underlying difficulty structure.

\begin{table}[t]
    \caption{Average provider-reported input tokens for compact edge-list, JSON, and PNML \citep{pnml} serializations.}
    \label{tab:serialization_length}
    \centering
    \small
    \setlength{\tabcolsep}{5pt}
    \begin{tabular}{@{}l|cccc@{}}
        \toprule
        Serialization & Easy & Medium & Hard & Overall $\downarrow$ \\
        \midrule
        Edge List & $2352$ & $4853$ & $8112$ & $5105$ \\
        JSON & $3362$ & $7115$ & $11963$ & $7480$ \\
        PNML & $8578$ & $18663$ & $31484$ & $19575$ \\
        \bottomrule
    \end{tabular}
\end{table}

\subsection{Prompting Technique}\label{app:prompting}

\begin{figure*}[t]
    \centering
    \includegraphics[width=1.0\textwidth]{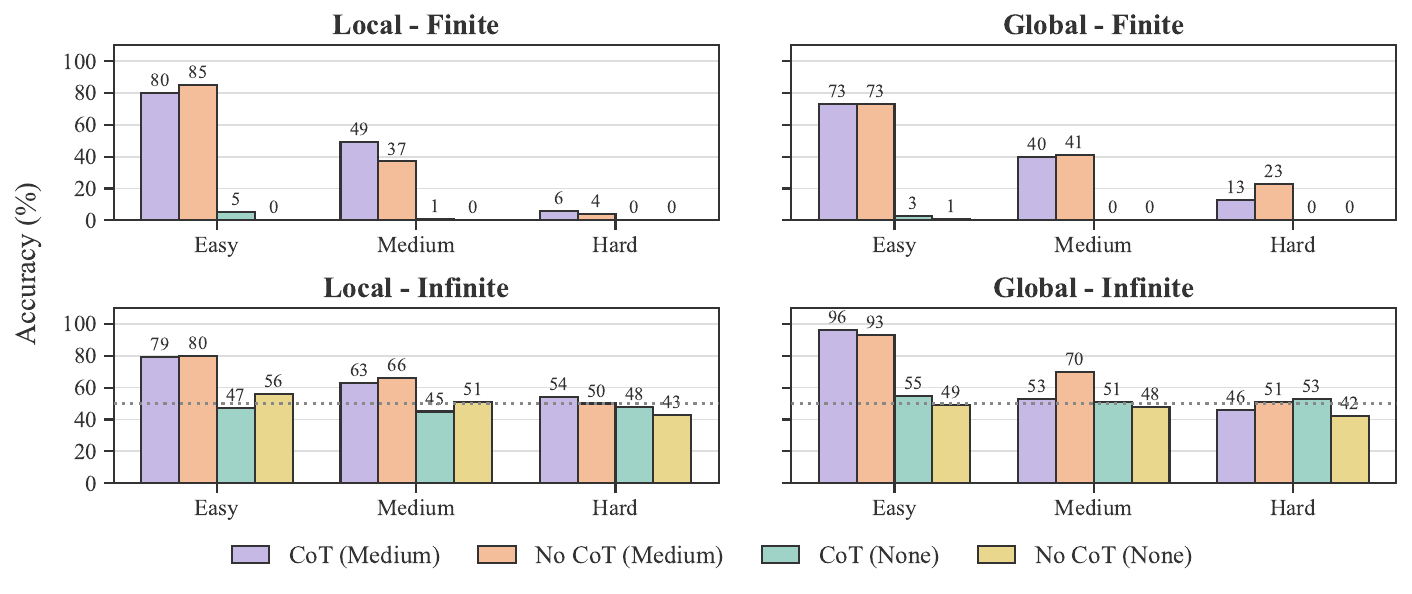}
    \caption{Exact-match accuracy of GPT-5.6 Sol with and without Chain-of-Thought (CoT) prompting under \texttt{Medium} and \texttt{None} reasoning effort, evaluated on the same fixed randomly sampled quarter of PetriBench. Results highlight that explicit CoT prompting has little effect at fixed reasoning effort, while disabling internal reasoning substantially degrades performance.}
    \label{fig:cot}
\end{figure*}

In line with common practice for evaluating multi-step reasoning, our main experiments use Chain-of-Thought (CoT) prompting \citep{wei2023chainofthoughtpromptingelicitsreasoning, kojima2023largelanguagemodelszeroshot}.
To assess the dependence of our results on this choice, we compare otherwise identical prompts with and without an explicit CoT instruction, with the complete prompt templates provided in Appendix~\ref{app:prompts}.
We perform this ablation using GPT-5.6 Sol on the same fixed, randomly sampled quarter of the benchmark used for the serialization study in Appendix~\ref{app:serialization}, under both \texttt{Medium} and \texttt{None} reasoning effort.
Figure~\ref{fig:cot} shows that CoT prompting has no consistent effect when reasoning effort is held fixed.
Paired bootstrap analysis further finds no statistically significant aggregate difference between CoT and no-CoT prompting under either reasoning setting, as reported in Table~\ref{tab:cot_significance}.

However, a substantially different picture emerges when reasoning effort itself is varied.
Across task families, disabling internal reasoning causes performance to collapse toward the benchmark baseline and largely removes the characteristic easy-to-hard performance gradient.
Most notably, CoT prompting does not compensate for this loss.
Even with internal reasoning disabled, the model can produce lengthy visible reasoning traces, in some cases comparable to or longer than the reasoning traces produced with internal reasoning enabled, without a corresponding improvement in accuracy.
Qualitative inspection reveals a recurring failure mode in which these traces assert intermediate state counts or structural conclusions without adequately deriving or verifying them, often relying on large unsupported assumptions that propagate to the final answer.
Together, these results indicate that eliciting a visible derivation is not itself sufficient for reliable state-space reasoning, while the model's internal test-time reasoning process is critical to performance.

\begin{table}[t]
    \caption{Aggregate effect of removing explicit CoT prompting. Confidence intervals are obtained from $10000$ paired bootstrap resamples over identical questions.}
    \label{tab:cot_significance}
    \centering
    \small
    \setlength{\tabcolsep}{5pt}
    \begin{tabular}{@{}l|ccc@{}}
        \toprule
        Reasoning Effort & CoT & No CoT & $\Delta$ [95\% CI] \\
        \midrule
        Medium & $54.3$ & $56.1$ & $+1.75$ [$-1.33$, $+4.83$] \\
        None   & $25.7$ & $24.2$ & $-1.50$ [$-3.75$, $+0.67$] \\
        \bottomrule
    \end{tabular}
\end{table}

\subsection{Shortcut Baselines}\label{app:shortcut_baselines}

Any procedural generation procedure necessarily induces a particular distribution over instances, making it possible in principle to recover aspects of the generation process rather than solve the intended reasoning problem.
Since covering the full distribution of Petri nets is neither practical nor necessarily desirable, our goal is instead to ensure that simple surface-level cues do not provide reliable shortcuts to the target labels.
We therefore test whether PetriBench can be solved from quantities directly observable in the serialized net using a shortcut baseline based on simple aggregate features.
The feature set includes counts of places, transitions, arcs, initial tokens, marked and zero-token places, minimum and maximum initial token counts, and multiplicity-$1$ and multiplicity-$2$ arcs.
For integer-valued tasks, we additionally include constants explicitly provided in the question, such as the firing horizon, target threshold, and initial tokens in the queried place.

A separate $L_2$-regularized linear model is fit for each question type, using logistic regression for binary tasks and ridge regression for integer-valued tasks, with the latter rounded to the nearest integer for exact-match evaluation.
We report five-fold cross-validation accuracy, with regularization selected independently within each training fold using nested cross-validation.
Table~\ref{tab:shortcut_baseline} shows that directly observable statistics provide little predictive signal beyond the corresponding trivial baselines.
Across all six tasks, the resulting change in accuracy remains within approximately five percentage points, with the shortcut predictor performing below baseline on the \texttt{Boundedness} and \texttt{Deadlock} tasks.
These results provide evidence that benchmark answers cannot be recovered from basic size, marking, and arc-multiplicity statistics alone.

\begin{table}[t]
    \caption{Accuracy gain of the shortcut baseline over the corresponding trivial baseline, using directly observable net and question statistics.}
    \label{tab:shortcut_baseline}
    \centering
    \small
    \setlength{\tabcolsep}{6pt}
    \begin{tabular}{@{}lc@{}}
        \toprule
        Task & $\Delta$ Accuracy (\%) \\
        \midrule
        Boundedness         & $-3.83$ \\
        Deadlock            & $-4.83$ \\
        $L_0$ Liveness      & $+4.00$ \\
        $L_4$ Liveness      & $+2.50$ \\
        Reachable Markings  & $+0.08$ \\
        Minimum Token Steps & $+1.08$ \\
        \bottomrule
    \end{tabular}
\end{table}

\subsection{Benchmark Correlation}\label{app:benchmark_correlation}

To assess whether PetriBench captures model capability in a manner consistent with established evaluations, we compare model rankings against the Artificial Analysis Intelligence, Agentic, and Long Context Reasoning indices \citep{artificialanalysis}.
We report Spearman rank correlations for the aggregate PetriBench score as well as each of the four taxonomy categories, using model configurations available in both evaluations.

\begin{figure*}[t]
    \centering
    \includegraphics[width=1.0\textwidth]{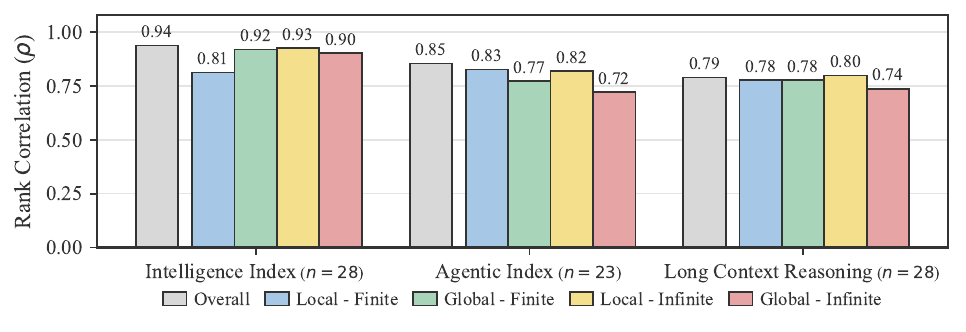}
    \caption{
    Spearman rank correlation between PetriBench model rankings and the Artificial Analysis Intelligence, Agentic, and Long Context Reasoning indices.
    Correlations are reported for aggregate PetriBench performance and each taxonomy category using model configurations shared between evaluations.
    PetriBench aligns very strongly with the broad Intelligence and Agentic indices, with somewhat lower agreement on Long Context Reasoning.
    }
    \label{fig:rank_correlations}
\end{figure*}

Figure~\ref{fig:rank_correlations} shows that PetriBench model rankings align strongly with external benchmarks.
The aggregate PetriBench ranking correlates most strongly with the Intelligence Index, followed closely by the Agentic Index, while the relationship with Long Context Reasoning is weaker but remains substantial.
This provides evidence that PetriBench captures a broad component of model reasoning capability rather than inducing an idiosyncratic ordering specific to the benchmark.
The lower correlation with Long Context Reasoning indicates that PetriBench is less closely aligned with long-context capability than with broader reasoning ability.

Figure~\ref{fig:rank_correlations} also shows that task-specific rankings remain strongly correlated with the same external indices.
This aligns with the analysis in Table~\ref{tab:difficulty_decomposition} and suggests that, although higher difficulty levels reveal more pronounced task-specific differences in model capability, the tasks still share a substantial common reasoning component rather than representing fully orthogonal abilities.

Together, the strong agreement with external reasoning benchmarks provides convergent evidence that PetriBench captures broad reasoning capability.
Importantly, it does so within a single compact, fully self-contained, and scalable formalism that is widely used to model real-world systems, without requiring domain-specific background knowledge, manually curated task creation, or open-ended judging.

\subsection{Error Analysis}\label{app:error_analysis}

To characterize model failures beyond aggregate accuracy, we conduct a structured qualitative analysis of incorrect reasoning traces, excluding parse failures.
We organize the analysis across model configurations, difficulty levels, tasks, and answer types, selecting up to three failures for each combination. 
When more than three failures are available, we select the shortest, median, and longest reasoning traces to capture a range of response lengths.
This procedure yields a corpus of $2{,}960$ failures across $36$ model configurations.

To avoid imposing a predefined taxonomy that could reflect human expectations or selective attention, each selected failure is first independently diagnosed by an OpenAI Codex agent \citep{openai_codex} using GPT-5.6 Sol at XHigh reasoning effort.
The agent receives the original prompt, model response, Petri net, alternative serializations, access to the TINA solver \citep{berthomieu2004tool}, and is instructed only to identify and describe the earliest concrete point at which the reasoning fails.
In a second stage, a single GPT-5.6 Sol XHigh agent reviews the resulting case-level diagnoses and produces an unconstrained set of fine-grained failure modes without reference to model identity, task, difficulty, or aggregate statistics.
These failure modes are subsequently analyzed and consolidated by a human into a compact set of minimally overlapping categories suitable for cross-task analysis.

The final taxonomy comprises five major failure modes.
\textbf{Transition Semantics} captures failures to correctly represent or execute the supplied net, including omitted or misread arcs, incorrect enabledness, mishandled token flow, and incorrect markings.
\textbf{Behavior Search} captures cases in which local transition semantics are handled correctly, but the reasoning focuses on an insufficient or irrelevant subset of the net and fails to explore a behavior needed to determine the answer, such as a counterexample or shorter route.
\textbf{Property Inference} captures invalid conclusions drawn from otherwise plausible local facts or behaviors, including incorrect invariants or insufficient justification for global properties.
\textbf{Solution Aggregation} covers errors in counting, deduplicating, comparing, or minimization.
Finally, \textbf{Answer Emission} is reserved for cases in which the substantive reasoning supports the correct conclusion but the reported Boolean or integer answer is incorrect.

As a single response may contain multiple downstream errors, we assign exactly one category corresponding to the earliest evidenced failure, while allowing no category when the evidence is insufficient to support a reliable classification.
After finalizing the taxonomy, all $2{,}960$ cases are independently reclassified from scratch by a fresh GPT-5.6 Sol XHigh agent with no access to the earlier category assignments.

\begin{figure*}[t]
    \centering
    \includegraphics[width=1.0\textwidth]{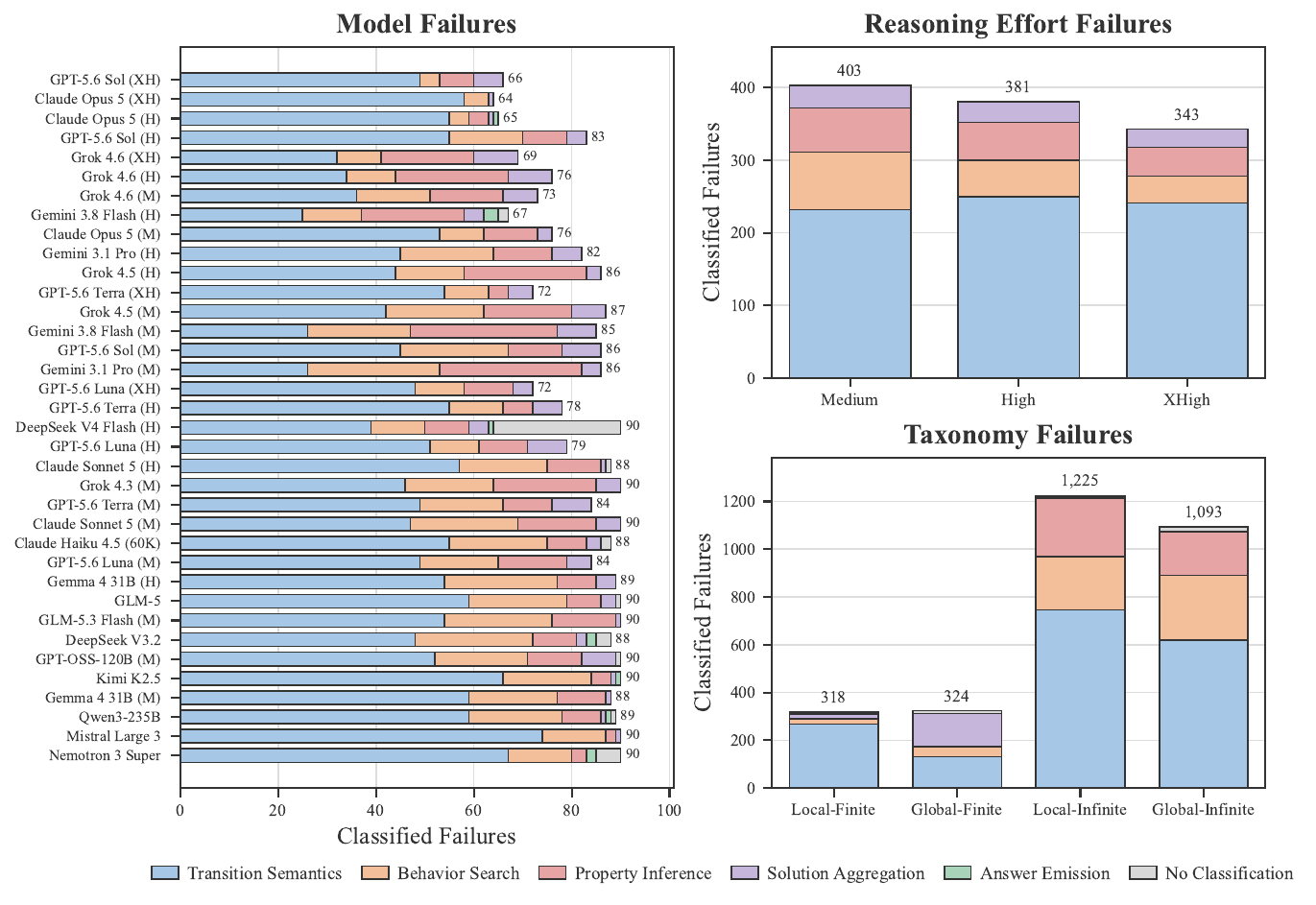}
    \caption{
    Distribution of classified PetriBench failure modes across models, reasoning-effort levels, and tasks.
    Categories correspond to the five-family taxonomy defined in Appendix~\ref{app:error_analysis}.
    Cases without sufficient evidence for a reliable assignment are shown as having no family.
    }
    \label{fig:failure_taxonomy}
\end{figure*}

Figure~\ref{fig:failure_taxonomy} summarizes the classified failure modes across individual model configurations, reasoning-effort levels for models that expose this control, and PetriBench task families.

Most notably, the reasoning-effort breakdown shows that the number of \textbf{Transition Semantics} failures remains comparatively stable as effort increases, whereas \textbf{Behavior Search}, \textbf{Property Inference}, and \textbf{Solution Aggregation} failures decrease.
Within the limitations of this qualitative sample, this suggests that additional reasoning effort does not simply improve low-level bookkeeping, but is associated more strongly with reductions in higher-level search and reasoning.

The taxonomy-level breakdown is also consistent with the structure of the underlying problems.
\texttt{ReachableMarkings}, which requires breadth-oriented state enumeration and deduplication, exhibits a substantially larger share of \textbf{Solution Aggregation} failures, whereas \texttt{MinimumTokenSteps}, which requires following valid firing sequences while tracking the evolving marking, is dominated by \textbf{Transition Semantics} errors.

Although the absence of \textbf{Property Inference} failures in the two finite-horizon tasks provides strong evidence for the validity of the semi-automated classification procedure (since these tasks do not require the global property reasoning captured by that category), we reiterate that labels are derived primarily through model-assisted trace analysis at a scale that precludes exhaustive human verification. 

\subsection{Further Difficulty Scaling}\label{app:difficulty_scaling}

\begin{figure*}[t]
    \centering
    \includegraphics[width=1.0\textwidth]{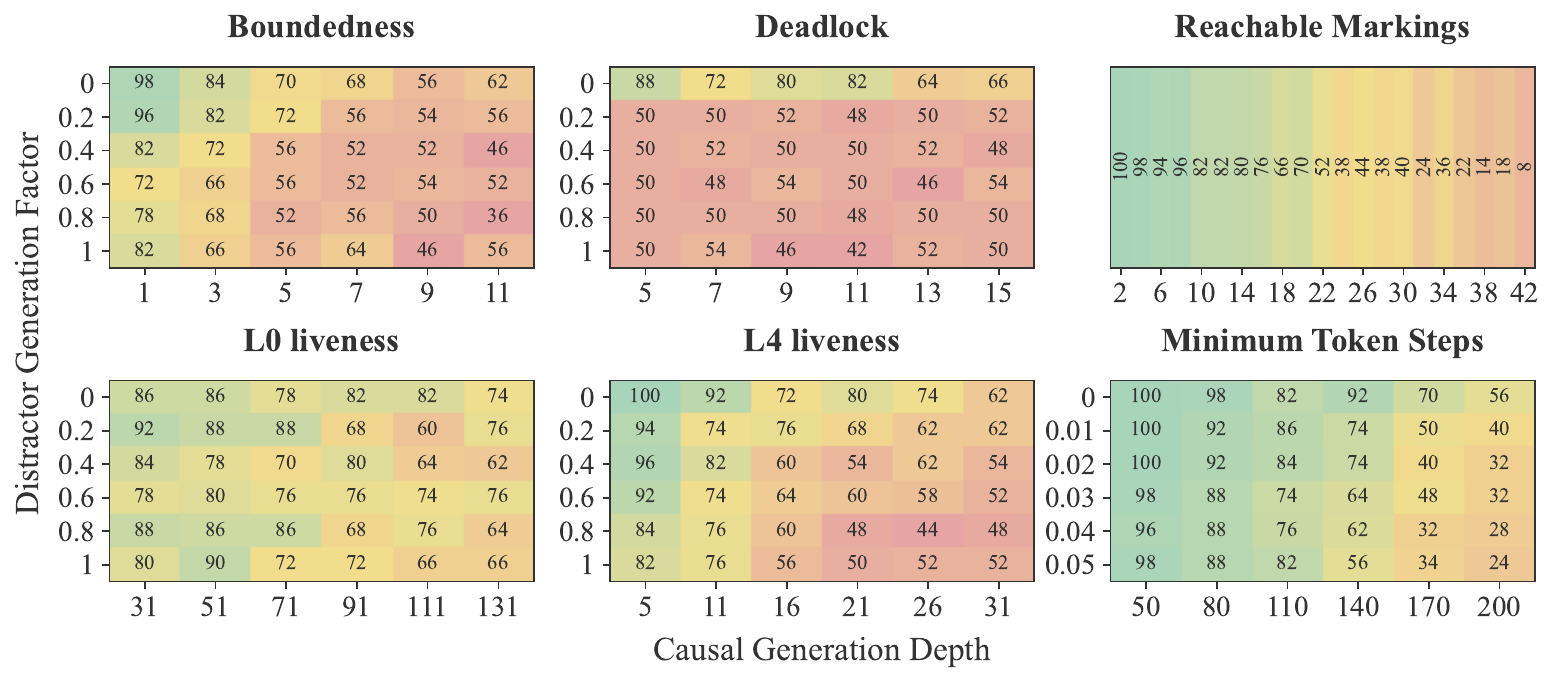}
    \caption{
    Controlled difficulty scaling for Gemini~3.8 Flash across PetriBench tasks.
    Each cell reports exact-match accuracy over $n=50$ instances as a function of Causal Generation Depth and Distractor Generation Factor.
    The same generation controls and parameter ranges as Figure~\ref{fig:difficulty_scaling} are used, enabling direct comparison of model-specific sensitivity to causal and distractor complexity.
    }
    \label{fig:gemini_scaling}
\end{figure*}

To complement the single-model structural scaling trends observed for GPT-5.6 Sol (Medium) in Section~\ref{ssec:difficulty_scaling}, we repeat the same controlled scaling experiment with Gemini~3.8 Flash (Medium) and $n=50$ questions per cell.
Figure~\ref{fig:gemini_scaling} varies \emph{Causal Generation Depth} together with the \emph{Distractor Generation Factor}, as defined in Appendix~\ref{app:generation}.

Notably, the two models respond strikingly differently to the same structural perturbations, revealing model-specific sensitivity to how difficulty is introduced rather than merely to its overall magnitude.
Additional distractor generation has a dramatically larger effect on Gemini~3.8 Flash for \texttt{Deadlock}, while \texttt{MinimumTokenSteps} is far more robust to distractors and remains dominated by causal depth, in sharp contrast to GPT-5.6 Sol.
Taken together, these results show that PetriBench difficulty is not only controllable but diagnostically rich, with the same generation controls producing systematic degradation across model families while nonetheless exposing different sensitivities.

\subsection{Further PetriBench Results}\label{app:petribench_model_diagnostics}

\begin{figure*}[t]
    \centering
    \includegraphics[width=1.0\textwidth]{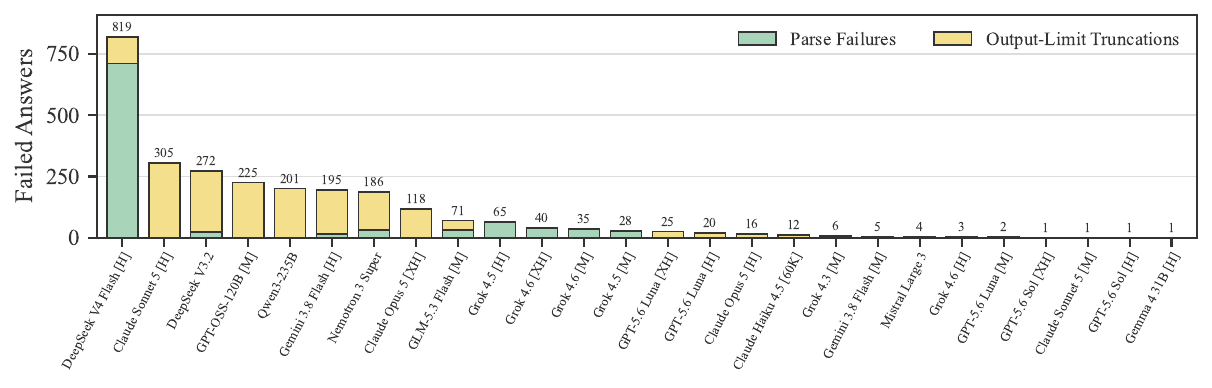}
    \caption{
    Failed responses by model configuration, separated into parse failures and output-limit truncations.
    Parse failures denote responses from which no valid final answer can be extracted under the deterministic rules described in Appendix~\ref{app:parsing}, while output-limit truncations denote responses terminated upon reaching the configured generation limit.
    Values are aggregated across tasks and difficulty levels.
    }
    \label{fig:parse_failures}
\end{figure*}

Under the deterministic parsing rules described in Appendix~\ref{app:parsing}, some model failures arise not only from incorrect reasoning but also from failures to produce an answer in a parseable form, reflecting issues with output formatting and instruction following.
Figure~\ref{fig:parse_failures} summarizes these cases by model configuration, separating parse failures from responses truncated at the output-token limit.
DeepSeek V4 Flash is a pronounced outlier, exhibiting substantially more failed responses than any other model, with the majority attributable to outputs that do not yield a parseable final answer under our deterministic extraction rules.

\begin{figure*}[t]
    \centering
    \includegraphics[width=1.0\textwidth]{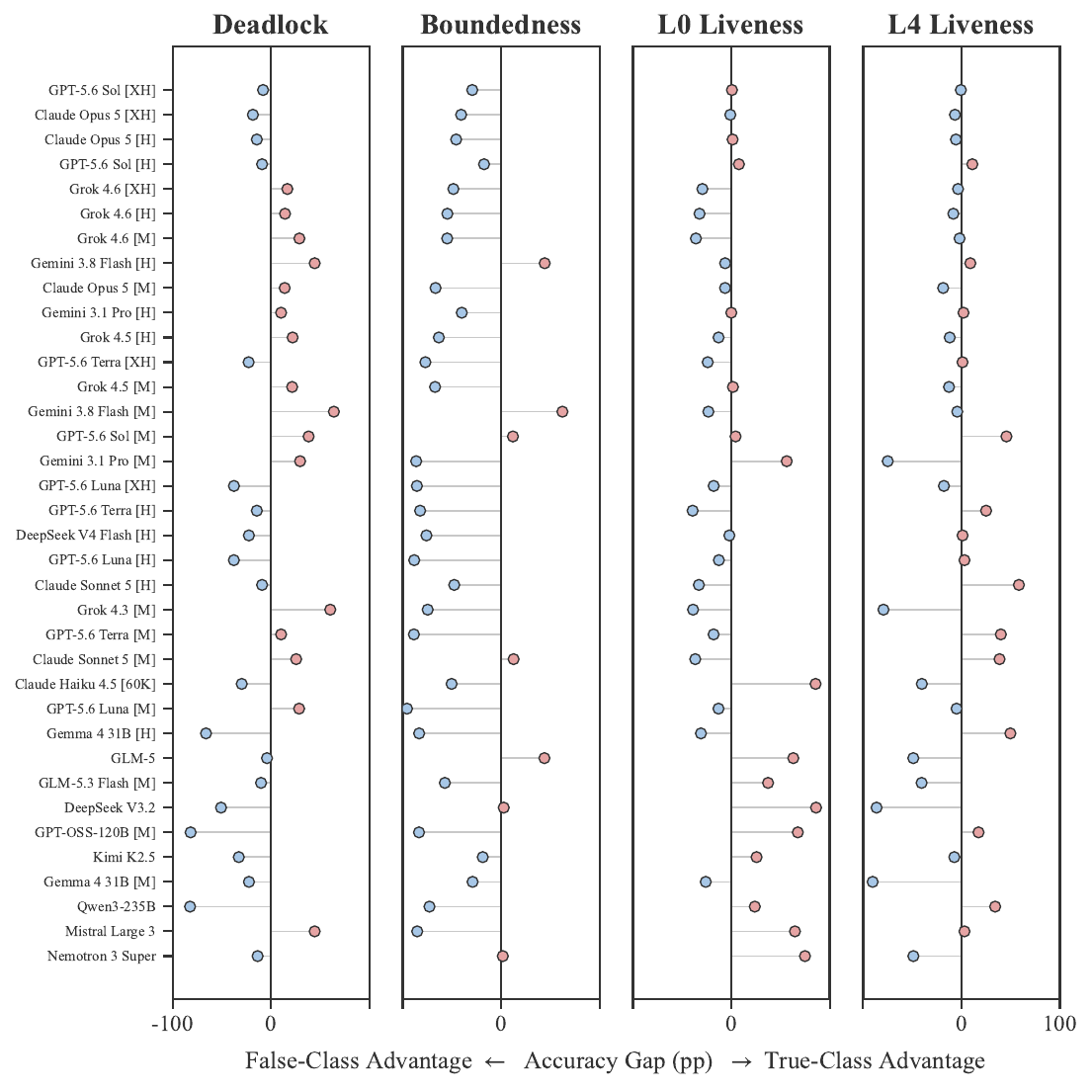}
    \caption{
    Label-conditioned accuracy differences on the Boolean PetriBench tasks.
    Each point reports $\mathrm{Acc}(y=\mathrm{True})-\mathrm{Acc}(y=\mathrm{False})$ for a model configuration, averaged equally across difficulty levels.
    Positive values indicate a True-class advantage and negative values a False-class advantage.
    }
    \label{fig:boolean_answer_bias}
\end{figure*}

Figure~\ref{fig:boolean_answer_bias} examines label-conditioned performance on the four Boolean PetriBench tasks.
To separate answer-selection behavior from formatting failures, we compute the gap only over responses from which a valid Boolean prediction can be parsed, while weighting True- and False-labeled instances equally across difficulty levels.
Positive values indicate relatively stronger performance on True-labeled instances, whereas negative values indicate relatively stronger performance on False-labeled instances.
The resulting patterns are strongly task dependent. Boundedness exhibits the clearest systematic asymmetry, with most model configurations performing substantially better on False instances, whereas L0 liveness shows the opposite tendency for several weaker models.
Deadlock and L4 liveness display more heterogeneous behavior across model families.
Stronger configurations are generally closer to balanced performance on several tasks, although substantial label-specific gaps remain for some models.

\begin{figure*}[t]
    \centering
    \includegraphics[width=1.0\textwidth]{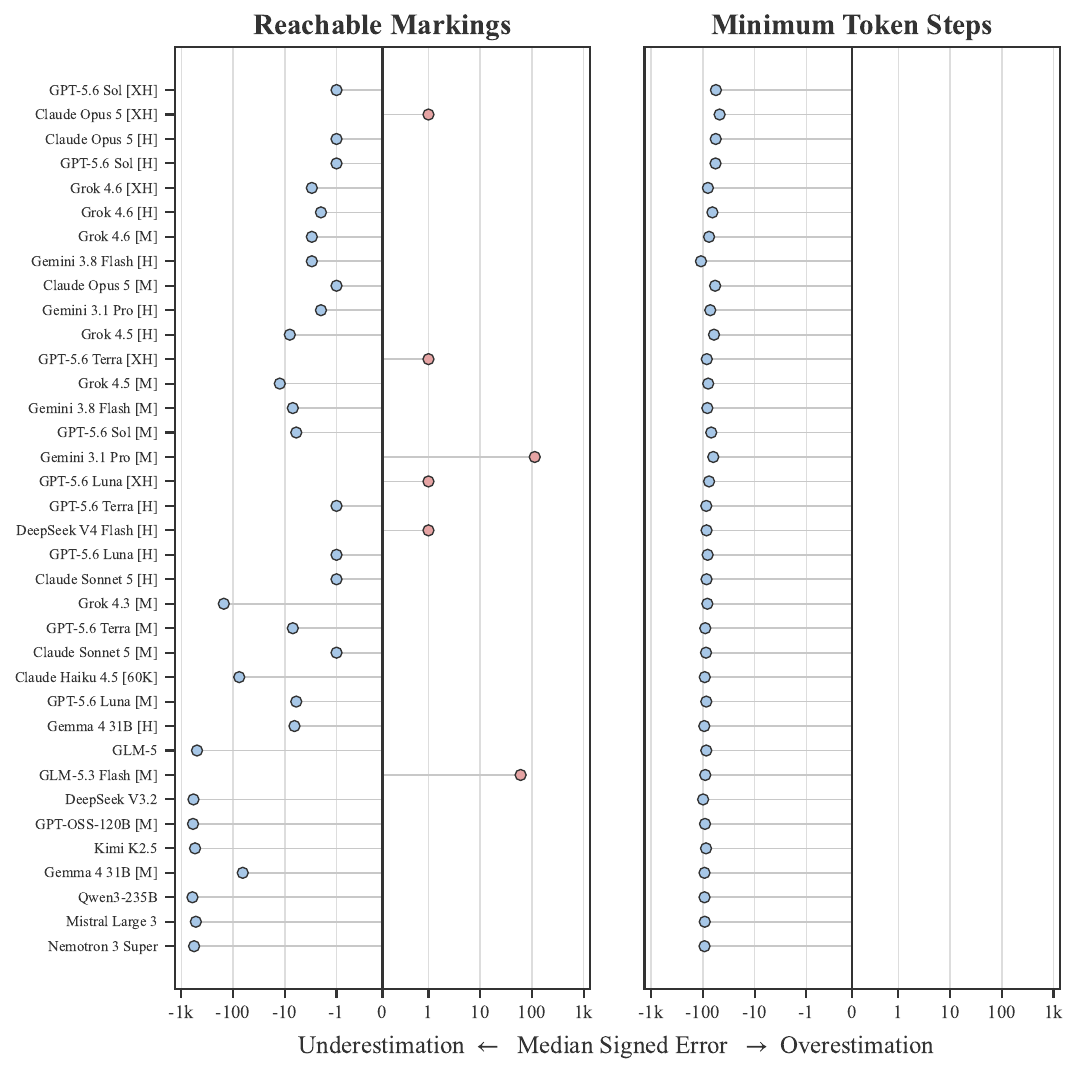}
    \caption{
    Median signed prediction error for the two integer-valued PetriBench tasks, computed over parseable responses and averaged across difficulty levels.
    Negative values indicate underestimation and positive values overestimation. \texttt{MinimumTokenSteps} shows a strong and consistent tendency toward underestimation, while \texttt{ReachableMarkings} exhibits a weaker but still prevalent underestimation pattern.
    }
    \label{fig:integer_error}
\end{figure*}

For the two integer-valued tasks, we additionally examine the direction and magnitude of prediction errors among parseable responses.
Figure~\ref{fig:integer_error} reports the median signed error for each model configuration, with negative values indicating underestimation and positive values indicating overestimation.
\texttt{MinimumTokenSteps} exhibits a pronounced and highly consistent underestimation bias across models, while \texttt{ReachableMarkings} also shows a broad tendency toward underestimation, albeit with greater variation across configurations.
We report the median rather than the mean to provide a robust summary that is not dominated by occasional large-magnitude errors from DeepSeek V4 Flash.

\begin{figure*}[t]
    \centering
    \includegraphics[width=1.0\textwidth]{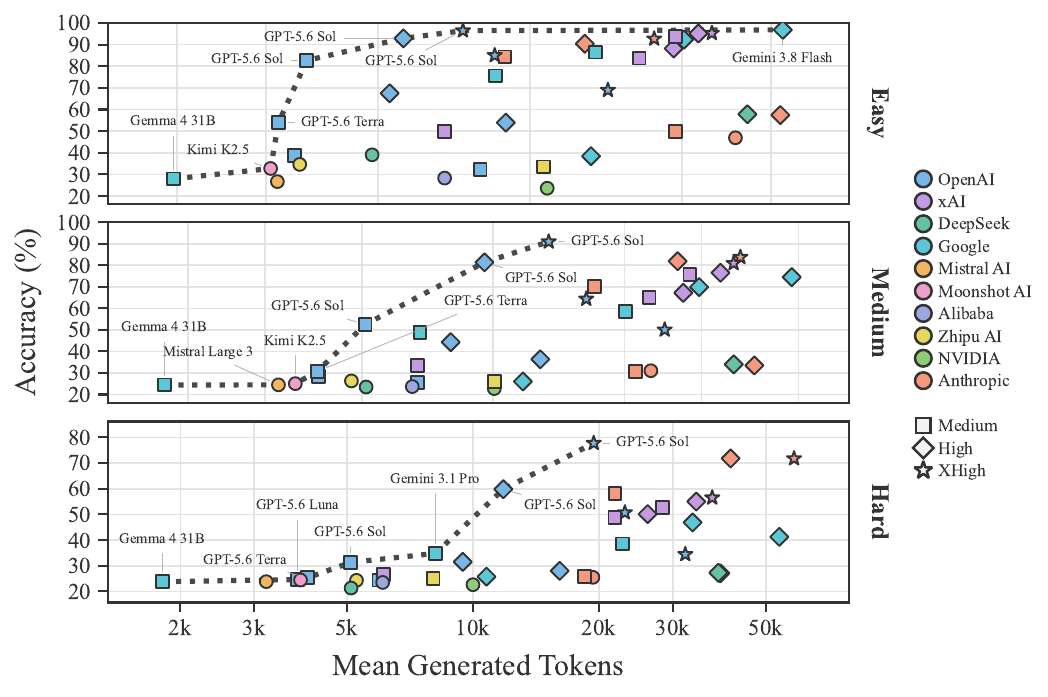}
    \caption{
    Accuracy versus reasoning-token Pareto frontiers across PetriBench difficulty levels.
    Each panel reports accuracy at a single difficulty level against the mean number of generated reasoning tokens for each model configuration.
    }
    \label{fig:pareto_difficulty}
\end{figure*}

\begin{figure*}[t]
    \centering
    \includegraphics[width=1.0\textwidth]{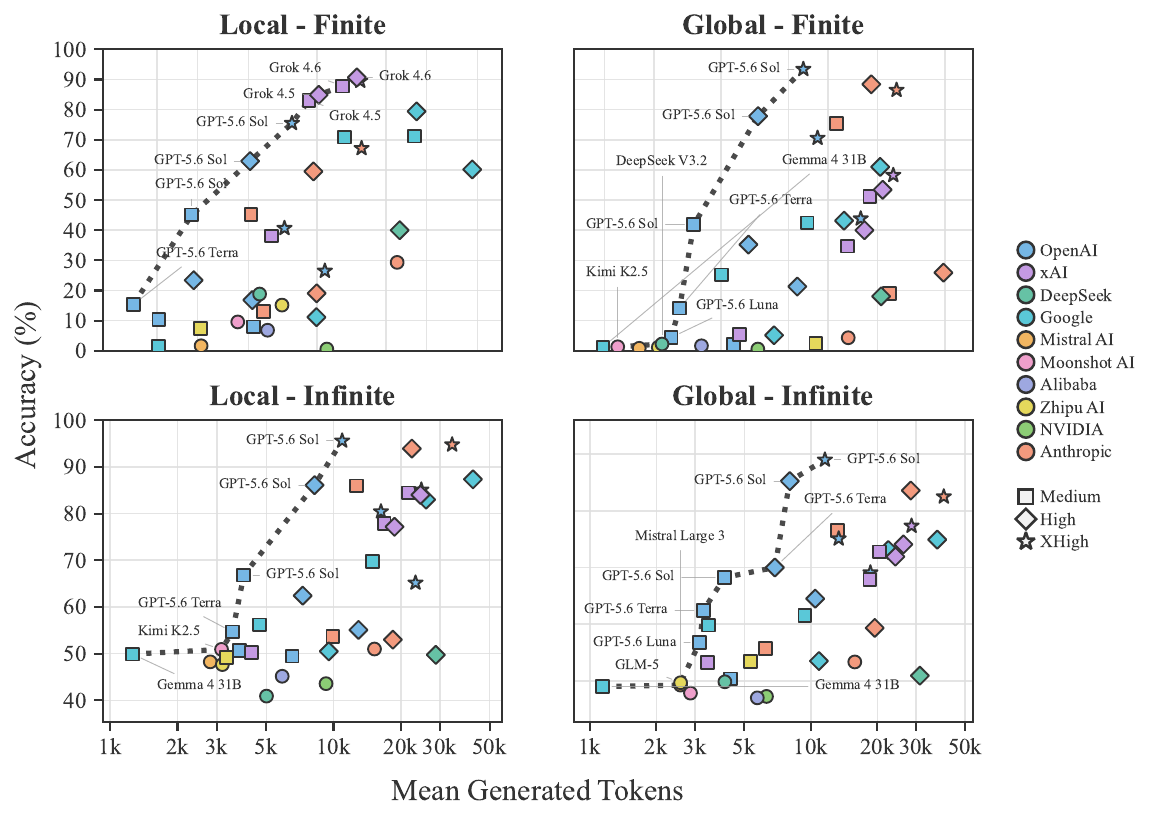}
    \caption{
    Accuracy versus reasoning-token Pareto frontiers across the four PetriBench taxonomy categories.
    Accuracy is averaged equally across difficulty levels within each category, while the horizontal axis reports mean generated reasoning tokens.
    }
    \label{fig:pareto_taxonomy}
\end{figure*}

To complement the aggregate reasoning-efficiency analysis in Section~\ref{ssec:reasoning_efficiency}, we additionally disaggregate the accuracy--compute relationship by benchmark difficulty and by PetriBench taxonomy category.
Figures~\ref{fig:pareto_difficulty} and~\ref{fig:pareto_taxonomy} report accuracy against mean generated reasoning tokens for each model configuration, with accuracy evaluated separately within each difficulty level and averaged across difficulty levels within each taxonomy category, respectively.

\begin{figure*}[t]
    \centering
    \includegraphics[width=1.0\textwidth]{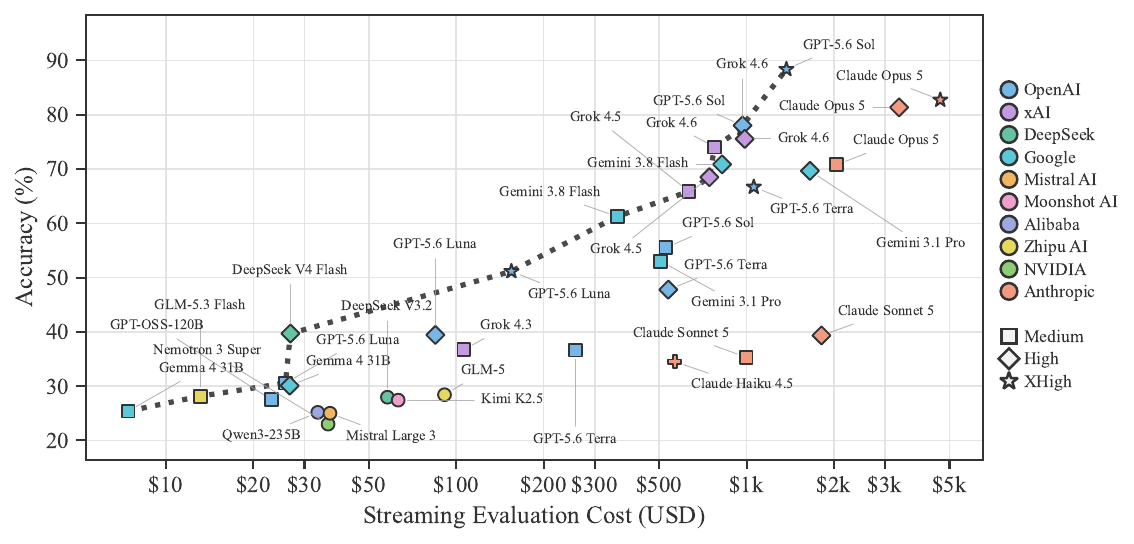}
    \caption{Accuracy versus cost Pareto frontier on PetriBench across evaluated model configurations.
    Cost is estimated using standard provider API pricing.}
    \label{fig:pareto_cost}
\end{figure*}

We additionally examine evaluation cost as a function of aggregate PetriBench accuracy.
Figure~\ref{fig:pareto_cost} reports estimated cost under standard provider API pricing and reveals a broader Pareto frontier than the reasoning-token analysis, with multiple model families occupying competitive regions of the cost--performance trade-off.
Under these rates, the complete set of reported PetriBench evaluations corresponds to approximately $26{,}373.05$
in inference cost, with actual costs reduced by batch discounts where supported.

Finally, Tables~\ref{tab:model_performance_boundedness}, \ref{tab:model_performance_deadlock}, \ref{tab:model_performance_reachable_markings}, \ref{tab:model_performance_min_token_steps}, \ref{tab:model_performance_l0_liveness}, and~\ref{tab:model_performance_l4_liveness} provide per-task breakdowns of model performance on PetriBench across all difficulty levels.

\begin{table*}[t]
    \caption{Exact-match accuracy (\%) on PetriBench for \texttt{Boundedness} across difficulty levels.
    Reasoning effort is indicated by M (Medium), H (High), and XH (XHigh). Best proprietary and open-weight results are highlighted in red and blue, respectively.}
    \label{tab:model_performance_boundedness}
    \centering
    \small
    \setlength{\tabcolsep}{3pt}
    \begin{tabular}{@{}lccc@{\hspace{1.5em}}lccc@{}}
        \toprule
        Model & Easy & Medium & Hard &
        Model & Easy & Medium & Hard \\
        \midrule
        Claude Opus 5 (XH) & $94.0$ & $74.0$ & $54.5$ &
        GPT-5.6 Luna (XH) & $71.5$ & $50.0$ & $50.0$ \\
        Claude Opus 5 (H) & $94.5$ & $74.5$ & $60.0$ &
        GPT-5.6 Luna (H) & $61.0$ & $50.5$ & $50.0$ \\
        Claude Opus 5 (M) & $80.5$ & $67.0$ & $52.5$ &
        GPT-5.6 Luna (M) & $51.5$ & $50.0$ & $50.0$ \\
        Grok 4.6 (XH) & $90.0$ & $66.5$ & $50.5$ &
        Grok 4.3 (M) & $57.0$ & $51.0$ & $49.0$ \\
        Grok 4.6 (H) & $90.0$ & $59.0$ & $51.0$ &
        Claude Sonnet 5 (H) & $61.0$ & $47.0$ & $52.5$ \\
        Grok 4.6 (M) & $93.0$ & $58.5$ & $49.0$ &
        Claude Sonnet 5 (M) & $54.0$ & $46.0$ & $58.0$ \\
        GPT-5.6 Sol (XH) & $94.5$ & \textcolor{softred}{$\bm{82.5}$} & \textcolor{softred}{$\bm{69.0}$} &
        Claude Haiku 4.5 & $54.5$ & $51.5$ & $50.5$ \\
        GPT-5.6 Sol (H) & $94.5$ & $80.5$ & $65.0$ &
        DeepSeek V4 Flash (H) & \textcolor{softblue}{$\bm{60.0}$} & $38.0$ & $40.5$ \\
        GPT-5.6 Sol (M) & $91.0$ & $57.5$ & $42.5$ &
        Gemma 4 31B (H) & $55.0$ & $52.0$ & $52.0$ \\
        Gemini 3.8 Flash (H) & \textcolor{softred}{$\bm{96.5}$} & $67.5$ & $55.5$ &
        Gemma 4 31B (M) & $46.0$ & $49.5$ & $50.0$ \\
        Gemini 3.8 Flash (M) & $69.5$ & $55.5$ & $51.5$ &
        GLM-5.3 Flash (M) & $50.5$ & $49.5$ & $48.0$ \\
        Grok 4.5 (H) & $87.5$ & $58.5$ & $49.0$ &
        GLM-5 & $50.5$ & $48.5$ & \textcolor{softblue}{$\bm{54.0}$} \\
        Grok 4.5 (M) & $75.0$ & $58.5$ & $49.5$ &
        DeepSeek V3.2 & $45.5$ & \textcolor{softblue}{$\bm{55.0}$} & $46.5$ \\
        GPT-5.6 Terra (XH) & $82.0$ & $52.5$ & $50.0$ &
        GPT-OSS-120B (M) & $53.5$ & $50.0$ & $50.0$ \\
        GPT-5.6 Terra (H) & $73.0$ & $50.5$ & $50.0$ &
        Kimi K2.5 & $54.0$ & $47.5$ & $52.5$ \\
        GPT-5.6 Terra (M) & $61.0$ & $50.0$ & $50.0$ &
        Qwen3-235B-A22B & $44.5$ & $45.5$ & $48.0$ \\
        Gemini 3.1 Pro (H) & $92.5$ & $57.0$ & $49.5$ &
        Mistral Large 3 & $49.0$ & $48.5$ & $48.5$ \\
        Gemini 3.1 Pro (M) & $60.0$ & $50.0$ & $49.5$ &
        Nemotron 3 Super & $46.5$ & $46.5$ & $46.5$ \\
        \bottomrule
    \end{tabular}
\end{table*}

\begin{table*}[t]
    \caption{Exact-match accuracy (\%) on PetriBench for \texttt{Deadlock} across difficulty levels.
    Reasoning effort is indicated by M (Medium), H (High), and XH (XHigh). Best proprietary and open-weight results are highlighted in red and blue, respectively.}
    \label{tab:model_performance_deadlock}
    \centering
    \small
    \setlength{\tabcolsep}{3pt}
    \begin{tabular}{@{}lccc@{\hspace{1.5em}}lccc@{}}
        \toprule
        Model & Easy & Medium & Hard &
        Model & Easy & Medium & Hard \\
        \midrule
        Claude Opus 5 (XH) & $98.5$ & $91.5$ & $82.5$ &
        GPT-5.6 Luna (XH) & $96.0$ & $83.0$ & $64.0$ \\
        Claude Opus 5 (H) & \textcolor{softred}{$\bm{100.0}$} & $95.0$ & $77.5$ &
        GPT-5.6 Luna (H) & $87.0$ & $77.5$ & $61.0$ \\
        Claude Opus 5 (M) & $99.5$ & $94.5$ & $65.0$ &
        GPT-5.6 Luna (M) & $81.5$ & $58.0$ & $49.5$ \\
        Grok 4.6 (XH) & $98.5$ & $96.5$ & $62.0$ &
        Grok 4.3 (M) & $63.0$ & $51.0$ & $48.5$ \\
        Grok 4.6 (H) & \textcolor{softred}{$\bm{100.0}$} & $90.0$ & $54.5$ &
        Claude Sonnet 5 (H) & $88.5$ & $59.0$ & $48.0$ \\
        Grok 4.6 (M) & $99.5$ & $88.0$ & $48.5$ &
        Claude Sonnet 5 (M) & $78.5$ & $50.0$ & $48.0$ \\
        GPT-5.6 Sol (XH) & $99.0$ & \textcolor{softred}{$\bm{97.0}$} & \textcolor{softred}{$\bm{92.0}$} &
        Claude Haiku 4.5 & $61.0$ & $51.0$ & $51.5$ \\
        GPT-5.6 Sol (H) & $97.5$ & $95.0$ & $79.0$ &
        DeepSeek V4 Flash (H) & \textcolor{softblue}{$\bm{67.0}$} & $52.0$ & $48.0$ \\
        GPT-5.6 Sol (M) & $99.0$ & $68.0$ & $51.5$ &
        Gemma 4 31B (H) & $59.0$ & $52.0$ & $51.0$ \\
        Gemini 3.8 Flash (H) & $99.5$ & $75.0$ & $55.5$ &
        Gemma 4 31B (M) & $58.5$ & $42.0$ & $48.0$ \\
        Gemini 3.8 Flash (M) & $92.5$ & $52.0$ & $48.5$ &
        GLM-5.3 Flash (M) & $66.0$ & \textcolor{softblue}{$\bm{54.5}$} & \textcolor{softblue}{$\bm{52.0}$} \\
        Grok 4.5 (H) & $91.0$ & $87.0$ & $58.5$ &
        GLM-5 & $49.0$ & $47.5$ & $48.5$ \\
        Grok 4.5 (M) & $83.5$ & $81.5$ & $59.0$ &
        DeepSeek V3.2 & $56.5$ & $49.5$ & $46.0$ \\
        GPT-5.6 Terra (XH) & $97.0$ & $92.0$ & $77.0$ &
        GPT-OSS-120B (M) & $51.5$ & $47.0$ & $50.0$ \\
        GPT-5.6 Terra (H) & $95.5$ & $87.5$ & $63.5$ &
        Kimi K2.5 & $47.0$ & $47.0$ & $39.0$ \\
        GPT-5.6 Terra (M) & $92.5$ & $68.0$ & $53.0$ &
        Qwen3-235B-A22B & $48.0$ & $46.0$ & $50.0$ \\
        Gemini 3.1 Pro (H) & \textcolor{softred}{$\bm{100.0}$} & $85.0$ & $54.5$ &
        Mistral Large 3 & $53.0$ & $51.0$ & $45.5$ \\
        Gemini 3.1 Pro (M) & $97.0$ & $50.5$ & $52.0$ &
        Nemotron 3 Super & $46.5$ & $49.5$ & $48.0$ \\
        \bottomrule
    \end{tabular}
\end{table*}

\newpage

\begin{table*}[t]
    \caption{Exact-match accuracy (\%) on PetriBench for \texttt{ReachableMarkings} across difficulty levels.
    Reasoning effort is indicated by M (Medium), H (High), and XH (XHigh). Best proprietary and open-weight results are highlighted in red and blue, respectively.}
    \label{tab:model_performance_reachable_markings}
    \centering
    \small
    \setlength{\tabcolsep}{3pt}
    \begin{tabular}{@{}lccc@{\hspace{1.5em}}lccc@{}}
        \toprule
        Model & Easy & Medium & Hard &
        Model & Easy & Medium & Hard \\
        \midrule
        Claude Opus 5 (XH) & $95.5$ & $91.8$ & $72.2$ &
        GPT-5.6 Luna (XH) & $64.8$ & $44.5$ & $22.2$ \\
        Claude Opus 5 (H) & $92.0$ & $91.2$ & $82.0$ &
        GPT-5.6 Luna (H) & $43.8$ & $16.8$ & $3.5$ \\
        Claude Opus 5 (M) & $86.8$ & $72.8$ & $67.0$ &
        GPT-5.6 Luna (M) & $11.2$ & $1.5$ & $0.5$ \\
        Grok 4.6 (XH) & $88.2$ & $59.5$ & $27.0$ &
        Grok 4.3 (M) & $15.0$ & $1.0$ & $0.0$ \\
        Grok 4.6 (H) & $87.5$ & $50.0$ & $22.8$ &
        Claude Sonnet 5 (H) & $51.0$ & $21.0$ & $5.8$ \\
        Grok 4.6 (M) & $82.2$ & $49.5$ & $21.8$ &
        Claude Sonnet 5 (M) & $39.3$ & $13.8$ & $2.5$ \\
        GPT-5.6 Sol (XH) & \textcolor{softred}{$\bm{98.2}$} & \textcolor{softred}{$\bm{95.8}$} & \textcolor{softred}{$\bm{86.2}$} &
        Claude Haiku 4.5 & $12.2$ & $0.8$ & $0.0$ \\
        GPT-5.6 Sol (H) & $94.0$ & $81.8$ & $57.8$ &
        DeepSeek V4 Flash (H) & \textcolor{softblue}{$\bm{34.2}$} & \textcolor{softblue}{$\bm{13.2}$} & \textcolor{softblue}{$\bm{6.8}$} \\
        GPT-5.6 Sol (M) & $75.0$ & $37.8$ & $12.8$ &
        Gemma 4 31B (H) & $13.8$ & $1.5$ & $0.2$ \\
        Gemini 3.8 Flash (H) & $91.0$ & $63.0$ & $29.0$ &
        Gemma 4 31B (M) & $3.5$ & $0.2$ & $0.0$ \\
        Gemini 3.8 Flash (M) & $80.8$ & $37.5$ & $9.0$ &
        GLM-5.3 Flash (M) & $5.8$ & $1.5$ & $0.2$ \\
        Grok 4.5 (H) & $70.5$ & $36.2$ & $13.2$ &
        GLM-5 & $3.0$ & $0.5$ & $0.0$ \\
        Grok 4.5 (M) & $66.5$ & $25.8$ & $12.0$ &
        DeepSeek V3.2 & $6.2$ & $0.5$ & $0.0$ \\
        GPT-5.6 Terra (XH) & $89.8$ & $71.5$ & $50.5$ &
        GPT-OSS-120B (M) & $6.0$ & $0.8$ & $0.0$ \\
        GPT-5.6 Terra (H) & $62.3$ & $32.2$ & $11.2$ &
        Kimi K2.5 & $3.5$ & $0.5$ & $0.0$ \\
        GPT-5.6 Terra (M) & $34.5$ & $6.8$ & $1.2$ &
        Qwen3-235B-A22B & $4.5$ & $0.5$ & $0.0$ \\
        Gemini 3.1 Pro (H) & $78.2$ & $37.8$ & $13.5$ &
        Mistral Large 3 & $2.5$ & $0.0$ & $0.0$ \\
        Gemini 3.1 Pro (M) & $59.2$ & $13.8$ & $2.5$ &
        Nemotron 3 Super & $1.2$ & $0.5$ & $0.0$ \\
        \bottomrule
    \end{tabular}
\end{table*}

\begin{table*}[t]
    \caption{Exact-match accuracy (\%) on PetriBench for \texttt{MinimumTokenSteps} across difficulty levels.
    Reasoning effort is indicated by M (Medium), H (High), and XH (XHigh). Best proprietary and open-weight results are highlighted in red and blue, respectively.}
    \label{tab:model_performance_min_token_steps}
    \centering
    \small
    \setlength{\tabcolsep}{3pt}
    \begin{tabular}{@{}lccc@{\hspace{1.5em}}lccc@{}}
        \toprule
        Model & Easy & Medium & Hard &
        Model & Easy & Medium & Hard \\
        \midrule
        Claude Opus 5 (XH) & $82.0$ & $64.8$ & $54.8$ &
        GPT-5.6 Luna (XH) & $48.5$ & $26.5$ & $4.5$ \\
        Claude Opus 5 (H) & $75.8$ & $56.2$ & $46.5$ &
        GPT-5.6 Luna (H) & $38.2$ & $11.0$ & $1.2$ \\
        Claude Opus 5 (M) & $70.2$ & $38.8$ & $26.8$ &
        GPT-5.6 Luna (M) & $26.5$ & $4.5$ & $0.0$ \\
        Grok 4.6 (XH) & $99.8$ & $94.8$ & $74.0$ &
        Grok 4.3 (M) & $77.0$ & $31.2$ & $6.0$ \\
        Grok 4.6 (H) & \textcolor{softred}{$\bm{100.0}$} & \textcolor{softred}{$\bm{95.2}$} & \textcolor{softred}{$\bm{76.8}$} &
        Claude Sonnet 5 (H) & $44.2$ & $12.2$ & $0.8$ \\
        Grok 4.6 (M) & $99.2$ & $93.2$ & $71.0$ &
        Claude Sonnet 5 (M) & $33.8$ & $5.0$ & $0.3$ \\
        GPT-5.6 Sol (XH) & $91.8$ & $81.5$ & $53.2$ &
        Claude Haiku 4.5 & $65.0$ & $22.8$ & $0.2$ \\
        GPT-5.6 Sol (H) & $86.8$ & $68.2$ & $33.8$ &
        DeepSeek V4 Flash (H) & \textcolor{softblue}{$\bm{72.5}$} & \textcolor{softblue}{$\bm{30.2}$} & \textcolor{softblue}{$\bm{17.2}$} \\
        GPT-5.6 Sol (M) & $77.5$ & $48.8$ & $9.0$ &
        Gemma 4 31B (H) & $33.5$ & $0.0$ & $0.0$ \\
        Gemini 3.8 Flash (H) & $98.5$ & $74.2$ & $7.8$ &
        Gemma 4 31B (M) & $4.2$ & $0.0$ & $0.0$ \\
        Gemini 3.8 Flash (M) & $99.2$ & $77.8$ & $36.8$ &
        GLM-5.3 Flash (M) & $19.8$ & $2.2$ & $0.2$ \\
        Grok 4.5 (H) & $97.5$ & $85.5$ & $71.8$ &
        GLM-5 & $39.2$ & $5.8$ & $0.5$ \\
        Grok 4.5 (M) & $96.0$ & $86.5$ & $66.8$ &
        DeepSeek V3.2 & $56.0$ & $0.5$ & $0.0$ \\
        GPT-5.6 Terra (XH) & $71.8$ & $31.5$ & $18.8$ &
        GPT-OSS-120B (M) & $23.8$ & $0.0$ & $0.0$ \\
        GPT-5.6 Terra (H) & $47.2$ & $17.8$ & $5.2$ &
        Kimi K2.5 & $27.8$ & $1.0$ & $0.0$ \\
        GPT-5.6 Terra (M) & $39.2$ & $6.0$ & $0.8$ &
        Qwen3-235B-A22B & $20.5$ & $0.0$ & $0.0$ \\
        Gemini 3.1 Pro (H) & $99.5$ & $88.8$ & $50.0$ &
        Mistral Large 3 & $5.0$ & $0.0$ & $0.0$ \\
        Gemini 3.1 Pro (M) & $98.0$ & $79.2$ & $35.5$ &
        Nemotron 3 Super & $1.8$ & $0.0$ & $0.0$ \\
        \bottomrule
    \end{tabular}
\end{table*}

\begin{table*}[t]
    \caption{Exact-match accuracy (\%) on PetriBench for \texttt{L0-Liveness} across difficulty levels.
    Reasoning effort is indicated by M (Medium), H (High), and XH (XHigh). Best proprietary and open-weight results are highlighted in red and blue, respectively.}
    \label{tab:model_performance_l0_liveness}
    \centering
    \small
    \setlength{\tabcolsep}{3pt}
    \begin{tabular}{@{}lccc@{\hspace{1.5em}}lccc@{}}
        \toprule
        Model & Easy & Medium & Hard &
        Model & Easy & Medium & Hard \\
        \midrule
        Claude Opus 5 (XH) & $97.5$ & $93.5$ & $90.5$ &
        GPT-5.6 Luna (XH) & $71.5$ & $51.5$ & $46.5$ \\
        Claude Opus 5 (H) & $94.5$ & $93.0$ & $86.5$ &
        GPT-5.6 Luna (H) & $56.5$ & $49.5$ & $49.0$ \\
        Claude Opus 5 (M) & $87.0$ & $84.5$ & $81.0$ &
        GPT-5.6 Luna (M) & $52.5$ & $52.0$ & $48.5$ \\
        Grok 4.6 (XH) & $99.5$ & $79.5$ & $58.5$ &
        Grok 4.3 (M) & $48.0$ & $51.5$ & $53.0$ \\
        Grok 4.6 (H) & $97.5$ & $78.5$ & $55.5$ &
        Claude Sonnet 5 (H) & $69.5$ & $49.5$ & $52.5$ \\
        Grok 4.6 (M) & $97.0$ & $80.5$ & $53.5$ &
        Claude Sonnet 5 (M) & $64.5$ & $53.5$ & $48.0$ \\
        GPT-5.6 Sol (XH) & $99.5$ & \textcolor{softred}{$\bm{97.0}$} & \textcolor{softred}{$\bm{91.0}$} &
        Claude Haiku 4.5 & $55.0$ & $50.0$ & $50.5$ \\
        GPT-5.6 Sol (H) & $94.5$ & $84.5$ & $77.0$ &
        DeepSeek V4 Flash (H) & \textcolor{softblue}{$\bm{62.5}$} & $42.5$ & $34.0$ \\
        GPT-5.6 Sol (M) & $82.5$ & $59.0$ & $55.0$ &
        Gemma 4 31B (H) & $51.5$ & $51.5$ & $48.5$ \\
        Gemini 3.8 Flash (H) & \textcolor{softred}{$\bm{100.0}$} & $83.0$ & $67.5$ &
        Gemma 4 31B (M) & $52.5$ & $51.5$ & $45.0$ \\
        Gemini 3.8 Flash (M) & $90.5$ & $73.0$ & $66.5$ &
        GLM-5.3 Flash (M) & $48.0$ & $50.5$ & $50.5$ \\
        Grok 4.5 (H) & $94.0$ & $64.5$ & $49.5$ &
        GLM-5 & $44.0$ & $52.0$ & $45.5$ \\
        Grok 4.5 (M) & $91.0$ & $65.0$ & $51.5$ &
        DeepSeek V3.2 & $39.0$ & $37.5$ & $39.0$ \\
        GPT-5.6 Terra (XH) & $85.5$ & $76.5$ & $63.0$ &
        GPT-OSS-120B (M) & $50.5$ & \textcolor{softblue}{$\bm{54.5}$} & $47.0$ \\
        GPT-5.6 Terra (H) & $75.5$ & $52.0$ & $49.5$ &
        Kimi K2.5 & $48.5$ & $54.0$ & \textcolor{softblue}{$\bm{51.0}$} \\
        GPT-5.6 Terra (M) & $65.0$ & $54.0$ & $45.0$ &
        Qwen3-235B-A22B & $41.5$ & $50.5$ & $44.5$ \\
        Gemini 3.1 Pro (H) & $93.5$ & $82.5$ & $64.0$ &
        Mistral Large 3 & $45.0$ & $50.0$ & $48.0$ \\
        Gemini 3.1 Pro (M) & $80.5$ & $55.5$ & $49.5$ &
        Nemotron 3 Super & $46.5$ & $44.0$ & $41.0$ \\
        \bottomrule
    \end{tabular}
\end{table*}

\begin{table*}[t]
    \caption{Exact-match accuracy (\%) on PetriBench for \texttt{L4-Liveness} across difficulty levels.
    Reasoning effort is indicated by M (Medium), H (High), and XH (XHigh). Best proprietary and open-weight results are highlighted in red and blue, respectively.}
    \label{tab:model_performance_l4_liveness}
    \centering
    \small
    \setlength{\tabcolsep}{3pt}
    \begin{tabular}{@{}lccc@{\hspace{1.5em}}lccc@{}}
        \toprule
        Model & Easy & Medium & Hard &
        Model & Easy & Medium & Hard \\
        \midrule
        Claude Opus 5 (XH) & $96.5$ & \textcolor{softred}{$\bm{98.0}$} & $92.5$ &
        GPT-5.6 Luna (XH) & $86.0$ & $74.0$ & $61.5$ \\
        Claude Opus 5 (H) & $98.5$ & $97.5$ & \textcolor{softred}{$\bm{93.5}$} &
        GPT-5.6 Luna (H) & $63.0$ & $58.0$ & $54.5$ \\
        Claude Opus 5 (M) & $93.5$ & $91.0$ & $78.5$ &
        GPT-5.6 Luna (M) & $49.0$ & $54.5$ & $48.0$ \\
        Grok 4.6 (XH) & $98.0$ & $96.0$ & $79.0$ &
        Grok 4.3 (M) & $48.0$ & $50.5$ & $50.5$ \\
        Grok 4.6 (H) & $98.5$ & $94.0$ & $80.0$ &
        Claude Sonnet 5 (H) & $50.0$ & $46.0$ & $50.5$ \\
        Grok 4.6 (M) & $98.5$ & $93.0$ & $84.0$ &
        Claude Sonnet 5 (M) & $54.0$ & $57.5$ & $44.5$ \\
        GPT-5.6 Sol (XH) & $98.5$ & $96.5$ & $91.0$ &
        Claude Haiku 4.5 & $50.5$ & $49.0$ & $51.0$ \\
        GPT-5.6 Sol (H) & $95.0$ & $91.0$ & $74.5$ &
        DeepSeek V4 Flash (H) & \textcolor{softblue}{$\bm{59.5}$} & \textcolor{softblue}{$\bm{52.5}$} & $47.5$ \\
        GPT-5.6 Sol (M) & $84.0$ & $62.5$ & $58.0$ &
        Gemma 4 31B (H) & $47.5$ & $50.0$ & \textcolor{softblue}{$\bm{54.0}$} \\
        Gemini 3.8 Flash (H) & \textcolor{softred}{$\bm{99.5}$} & $96.0$ & $78.0$ &
        Gemma 4 31B (M) & $51.0$ & $52.0$ & $47.5$ \\
        Gemini 3.8 Flash (M) & $80.0$ & $58.5$ & $50.0$ &
        GLM-5.3 Flash (M) & $52.5$ & $46.0$ & $48.0$ \\
        Grok 4.5 (H) & $97.0$ & $84.0$ & $74.0$ &
        GLM-5 & $49.0$ & $50.0$ & $45.5$ \\
        Grok 4.5 (M) & $95.5$ & $90.5$ & $73.5$ &
        DeepSeek V3.2 & $47.0$ & $44.0$ & $39.0$ \\
        GPT-5.6 Terra (XH) & $92.5$ & $88.0$ & $77.0$ &
        GPT-OSS-120B (M) & $44.5$ & $52.0$ & $48.5$ \\
        GPT-5.6 Terra (H) & $77.0$ & $64.5$ & $56.0$ &
        Kimi K2.5 & $50.0$ & $49.0$ & $53.0$ \\
        GPT-5.6 Terra (M) & $65.5$ & $48.0$ & $50.5$ &
        Qwen3-235B-A22B & $42.5$ & $46.5$ & $45.5$ \\
        Gemini 3.1 Pro (H) & $96.5$ & $81.5$ & $80.0$ &
        Mistral Large 3 & $51.5$ & $46.5$ & $48.5$ \\
        Gemini 3.1 Pro (M) & $52.0$ & $48.0$ & $51.5$ &
        Nemotron 3 Super & $43.5$ & $41.0$ & $45.5$ \\
        \bottomrule
    \end{tabular}
\end{table*}

\FloatBarrier

\section{Implementation Details}\label{app:implementation_details}
\subsection{Petri Net Generation}\label{app:generation}
\subsubsection{Overview and Principles}\label{app:generation_principle}
We generate weighted Petri nets $\mathcal{N}=(P,T,F,W,M_0)$, where every  arc multiplicity lies in $\{1,2\}$.
All generated nets are weakly connected when viewed as bipartite graphs over places and transitions.
Our task-specific generators are property preserving: each generator first constructs the places, transitions, and initial marking that determine the answer, which we refer to as the \emph{core}, and subsequently introduces additional paths, cycles, choices, and synchronization dependencies under restrictions that preserve the queried property.
This avoids repeatedly sampling arbitrary nets and invoking a solver until one with the desired label is found.
The exception is \texttt{ReachableMarkings}, for which instances are produced through randomized structural generation and the exact integer answer is subsequently computed with the TINA solver \citep{berthomieu2004tool}.
For all Petri Nets included in PetriBench, we additionally cross-check the stored labels using the TINA solver and task-specific formal proofs or counterexamples.

The generation procedure distinguishes answer-relevant structure from distractor structure introduced to increase the difficulty of isolating and reasoning over the answer-determining portion of the net.
Distractor structure remains integrated with the rest of the net through ordinary arcs, synchronization transitions, and auxiliary places with balanced token flow.
Complexity is therefore not increased by appending disconnected components.

Two principal controls govern structural complexity.
The \emph{causal generation depth}, denoted by $d$, controls the extent of the answer-determining construction.
For most tasks, this corresponds to the number of randomized expansions applied to the core, while for \texttt{MinimumTokenSteps} it directly determines the required minimum firing count.
The \emph{distractor generation factor}, denoted by $\rho$, controls the amount of additional label-preserving structure relative to the causal construction.
For each instance, $d$ is sampled uniformly from an inclusive range assigned to its difficulty level.

The number of additional distractor generation iteration steps is obtained by multiplying $\rho$ by an appropriate measure of the fully generated answer-relevant construction and rounding to the nearest integer.
This reference quantity is the number of answer-relevant transitions for \texttt{Boundedness} and both liveness questions, the number of resource-contending places for \texttt{Deadlock}, and the number of transitions needed to produce the required target tokens for \texttt{MinimumTokenSteps}.
Thus, the numerical values of $d$ and $\rho$ are task-specific and are not directly comparable across question types.

Table~\ref{tab:generation_values} reports the generation values used for PetriBench.
Each generator additionally exposes task-specific parameters that determine the baseline distribution of generated nets.
We hold these parameters fixed to isolate the effects of $d$ and $\rho$, since varying them does not generally induce a clean or monotonic change in reasoning difficulty.
For example, increasing the number of initial tokens need not increase difficulty when the underlying transition dependencies remain unchanged.
The fixed values were chosen empirically to elicit the consistent scaling trends observed in Figure~\ref{fig:difficulty_scaling}.

In Table~\ref{tab:generation_values}, $\ell$ denotes the size of the initial answer-relevant construction before iterative expansion.
The parameter $e$ denotes additional tokens, except for \texttt{Deadlock}, where it is implemented through additional resource-contending places.
For \texttt{MinimumTokenSteps}, $\delta$ controls the density of safe additional arcs within the distractor structure, while $k$ specifies the maximum firing horizon for \texttt{ReachableMarkings}.
As discussed in Section~\ref{ssec:pn_generation}, \texttt{ReachableMarkings} does not use a distractor factor because every additional fireable structure can alter the reachable-state count.

Consequently, the resulting net sizes are not prescribed directly, but emerge from the sampled generation parameters and the structures introduced by each generation operation.
Table~\ref{tab:realized-net-sizes} reports the resulting mean net sizes.

\begin{table*}[t]
  \centering
  \small
  \setlength{\tabcolsep}{4pt}
  \begin{tabular}{@{}lcccccccccc@{}}
      \toprule
      & \multicolumn{4}{c}{Fixed Parameters}
      & \multicolumn{2}{c}{Easy}
      & \multicolumn{2}{c}{Medium}
      & \multicolumn{2}{c}{Hard} \\
      \cmidrule(lr){2-5}
      \cmidrule(lr){6-7}
      \cmidrule(lr){8-9}
      \cmidrule(l){10-11}
      Task
          & $\ell$ & $e$ & $k$ & $\delta$
          & $d$ & $\rho$
          & $d$ & $\rho$
          & $d$ & $\rho$ \\
      \midrule
      Boundedness
          & $5$ & $24$ & -- & --
          & $[3,4]$   & $0.20$
          & $[7,8]$   & $0.60$
          & $[11,12]$ & $1.00$ \\
      Deadlock
          & $4$ & $24$ & -- & --
          & $[4,5]$   & $0$
          & $[8,9]$   & $0.40$
          & $[12,13]$ & $0.80$ \\
      Reachable Markings
          & $5$ & -- & $3$ & --
          & $[13,14]$ & --
          & $[27,28]$ & --
          & $[41,42]$ & -- \\
      Minimum Token Steps
          & -- & $100$ & -- & $0.75$
          & $[20,50]$   & $0.01$
          & $[80,110]$  & $0.03$
          & $[140,170]$ & $0.05$ \\
      $L_0$ Liveness
          & $16$ & $24$ & -- & --
          & $[50,51]$   & $0$
          & $[90,91]$   & $0.40$
          & $[130,131]$ & $0.80$ \\
      $L_4$ Liveness
          & $16$ & $24$ & -- & --
          & $[10,11]$ & $0.40$
          & $[20,21]$ & $0.60$
          & $[30,31]$ & $0.80$ \\
      \bottomrule
  \end{tabular}
  \caption{
      Generation values used for the creation of PetriBench.
      The interpretation of one relevant or irrelevant generation step depends on the queried property.
  }
  \label{tab:generation_values}
\end{table*}

\begin{table*}[t]
  \centering
  \small
  \setlength{\tabcolsep}{4.5pt}
  \begin{tabular}{@{}lccccccccc@{}}
      \toprule
      &
      \multicolumn{3}{c}{Places} &
      \multicolumn{3}{c}{Transitions} &
      \multicolumn{3}{c}{Arcs} \\
      \cmidrule(lr){2-4}
      \cmidrule(lr){5-7}
      \cmidrule(l){8-10}
      Task
      & Easy & Medium & Hard
      & Easy & Medium & Hard
      & Easy & Medium & Hard \\
      \midrule
      Boundedness
          & $77.0$ & $110.7$ & $144.0$
          & $74.2$ & $118.1$ & $161.3$
          & $217.4$ & $399.8$ & $631.7$ \\
      Deadlock
          & $34.2$ & $79.9$ & $146.7$
          & $25.6$ & $62.6$ & $116.2$
          & $85.4$ & $217.4$ & $412.9$ \\
      Reachable Markings
          & $20.5$ & $35.7$ & $51.0$
          & $23.4$ & $41.4$ & $59.7$
          & $72.8$ & $134.8$ & $197.9$ \\
      Minimum Token Steps
          & $52.7$ & $132.0$ & $218.2$
          & $45.9$ & $120.7$ & $199.3$
          & $146.1$ & $382.5$ & $639.8$ \\
      $L_0$ Liveness
          & $100.7$ & $174.7$ & $246.0$
          & $149.5$ & $350.7$ & $621.8$
          & $360.0$ & $922.1$ & $1733.1$ \\
      $L_4$ Liveness
          & $30.9$ & $47.1$ & $65.1$
          & $58.5$ & $104.1$ & $161.2$
          & $155.7$ & $282.1$ & $449.7$ \\
      \bottomrule
  \end{tabular}
  \caption{
      Mean realized net sizes in PetriBench, broken down by task.
  }
  \label{tab:realized-net-sizes}
\end{table*}

\subsubsection{Causal Generation Specifics by Task}\label{app:causal_details}

\paragraph{Minimum Token Steps}
For a minimum token task, the generator samples the answer $d$ from the configured causal-depth range and chooses a target token count $c\in\{1,2\}$.
The target place initially contains no tokens.
The generator then constructs the net so that the shortest firing sequence that places at least $c$ tokens in the target has length $d$.

The required tokens may reach the target through a long sequence of transitions, through several paths that must be completed and synchronized, or through a cycle that must execute several times before producing enough tokens.
Some instances also contain shorter-looking alternatives, but these either produce too few tokens, require an input that can never become marked, or circulate tokens without increasing the target marking.
The remaining transitions are constructed so that any other way of producing the required target tokens must use at least $d$ firings.

The density parameter $\delta$ adds additional fully randomized arcs between places and transitions among only the additional distractor structure, while extra initial tokens $e$ are placed only where they cannot create a shorter path to the target.

\paragraph{Reachable Markings}
Each reachable-marking instance is generated by starting from a cyclic net with five places and repeatedly applying randomly selected structural modifications.
A modification may replace a transition with a longer sequence, introduce alternative firing paths, split a path into concurrent branches that later rejoin, add a local cycle, or connect previously modified regions of the net.
Sequence lengths, branch widths, and cycle lengths are sampled from $\{2,3\}$.
New modifications are more likely to be applied near recently modified transitions, encouraging successive changes to interact and producing tightly connected structures rather than collections of independent branches.

The modified net is then analyzed with TINA \citep{berthomieu2004tool} to enumerate all markings reachable within the firing horizon, which is fixed to $k=3$ in the main benchmark.
Since any additional fireable structure can change this reachable set, all generated structure is treated as answer-relevant and no separate distractor generation factor is used for this task.

\paragraph{Liveness}
We generate liveness instances independently for $L_0 \in \{\mathrm{True}, \mathrm{False}\}$ and $L_4 \in \{\mathrm{True}, \mathrm{False}\}$, with each net constructed around maintaining the property for a single queried transition.

The construction begins with a cycle that circulates the tokens required to enable the queried transition.
As the causal generation depth increases, existing transitions are replaced or supplemented with longer sequences, alternative branches, loops, and additional paths whose tokens must be brought simultaneously to the input places of the queried transition.
Transitions that depend on several paths couple their evolution and make the required marking more difficult to identify.
Any additional initial tokens are confined to distractor structure of the net from which they cannot affect whether the queried transition can be enabled.

For an $L_0=\mathrm{True}$ instance, the queried transition requires tokens from two mutually exclusive places in the same cycle.
The available tokens cannot occupy both places simultaneously, so the transition can never fire.
For an $L_0=\mathrm{False}$ instance, the required tokens occur in cyclic subnets whose markings can be moved into a configuration that enables the
queried transition.
For an $L_4=\mathrm{True}$ instance, the tokens required by the queried transition can always be circulated back to its input places.
Thus, from every reachable marking, there is a firing sequence that enables the transition.
When the queried transition fires, it returns the same weighted number of
tokens to the corresponding cycles. Transitions that could permanently remove these tokens are made unreachable by requiring tokens from mutually exclusive places.
For an $L_4=\mathrm{False}$ instance, the net instead contains a reachable one-way transition that moves a required token into a closed set of acyclic places from which it cannot return.
Once this transition fires, the queried transition can never be enabled again.

\paragraph{Boundedness}
To generate a bounded net, we ensure that firing transitions can never increase a global weighted token count. Each place $p$ is assigned a positive integer weight $\alpha_p$, so that a token in that place contributes $\alpha_p$ units to the total
\[
      V(M)=\sum_{p\in P}\alpha_p M(p).
\]
Every transition is constructed so that the total weight of the tokens it produces is no greater than the total weight of the tokens it consumes.
Consequently, $V(M)$ can never exceed its finite initial value $V(M_0)$.
Because every place has positive weight, no individual place can accumulate arbitrarily many tokens, and the net is therefore bounded.
Additional paths, cycles, choices, and shortcuts are added only when they preserve this non-increasing weighted token count.

To generate an unbounded net, we construct a firing sequence that can be repeated indefinitely.
After one execution of this sequence, every place needed to execute it again contains at least as many tokens as before, while at least one place contains more tokens.
The same sequence can therefore be repeated, increasing the number of tokens on every repetition.
Additional paths and transitions are added in a way so they do not prevent this sequence from being executed.

Bounded and unbounded nets are generated in matched pairs with similar numbers of places, transitions, arcs, initial tokens, and arc multiplicities.
These aggregate properties therefore cannot be used as simple cues for predicting the answer.

\paragraph{Deadlock}
For a sampled depth $d$, the deadlock generator creates $\ell+d$ resource-contending subnets with shared resource places.
Each subnet has a start place, a waiting place, and a place representing possession of both required resources.
Its transitions acquire the two resources in separate firings and subsequently release them before returning to the start place.
The token and arc multiplicities used to represent a resource or a resource-contending unit are sampled from $\{1,2\}$.
The extra-token parameter adds further complete units to the start places instead of placing additional tokens in intermediate states.

As for boundedness tasks, deadlocking and non-deadlocking instances are generated with similar sizes, differing only in the assignment of shared resource places to the acquisition transitions.
In a deadlocking instance, these assignments form a cycle so that each subnet can acquire one resource while waiting for a resource held by the next subnet.
After every subnet acquires its first resource, all
resources are occupied and no second acquisition can occur, resulting in a marking with no enabled transitions.
In a non-deadlocking instance, the resources are instead assigned a random total order, and every subnet acquires its lower-ranked resource before its higher-ranked resource.
A cyclic dependency can therefore never form, meaning that at every reachable marking, at least one resource-acquisition or release transition remains enabled.

\subsection{Benchmark Construction}
\subsubsection{Composition}
As described in Section~\ref{ssec:benchmark_composition}, PetriBench is organized according to the four combinations of local or global scope and finite or infinite temporal extent introduced in Section~\ref{ssec:task_taxonomy}.
We assign the same number of questions to each of these four categories so that the aggregate score does not disproportionately reflect any one task type.
Within each category, the constituent tasks and labels are also balanced wherever the answer is Boolean.
The main benchmark contains $4{,}800$ questions divided equally among Easy, Medium, and Hard difficulty levels.
Each level therefore contains $1{,}600$ questions, namely $400$ \texttt{MinimumTokenSteps} questions, $400$ \texttt{ReachableMarkings} questions, $200$ \texttt{L0-Liveness} questions, $200$ \texttt{L4-Liveness} questions, $200$ \texttt{Deadlock} questions, and $200$ \texttt{Boundedness} questions.
Every generated net contributes exactly one question to the benchmark.
This prevents several closely related questions about the same net from being treated as independent observations and ensures that larger nets do not receive greater weight merely because they support more possible queries.
Table~\ref{tab:benchmark-composition} summarizes the resulting composition.

\begin{table*}[t]
    \centering
    \small
    \setlength{\tabcolsep}{9pt}
    \begin{tabular}{@{}lllccc@{}}
        \toprule
        Scope & Horizon & Task & Answer Type & Per Difficulty & Total \\
        \midrule
        Local  & Finite   & Minimum Token Steps & Integer & 400 & 1,200 \\
        \addlinespace[0.3em]
        Local  & Infinite & $L_0$ Liveness       & Boolean & 200 & 600 \\
        Local  & Infinite & $L_4$ Liveness       & Boolean & 200 & 600 \\
        \addlinespace[0.3em]
        Global & Finite   & Reachable Markings   & Integer & 400 & 1,200 \\ \addlinespace[0.3em]
        Global & Infinite & Deadlock              & Boolean & 200 & 600 \\
        Global & Infinite & Boundedness           & Boolean & 200 & 600 \\
        \bottomrule
    \end{tabular}
    \caption{
        Composition of the main PetriBench evaluation across the four categories of the task taxonomy and three difficulty levels.
    }
    \label{tab:benchmark-composition}
\end{table*}

\subsubsection{Serialization}

All Petri nets in the main benchmark are serialized using a compact edge-list format and inserted directly into the corresponding task prompt.
This format provides a concise textual description of the initial marking and the connections between places and transitions without attaching any additional semantic meaning to the elements of the net.

For example,
\begin{lstlisting}
place p0 2
place p1 0
trans t0
arc p0 t0 1
arc t0 p1 2
\end{lstlisting}
specifies two places and one transition: \texttt{p0} initially contains two tokens, and firing \texttt{t0} consumes one token from \texttt{p0} and produces two tokens in \texttt{p1}.
Before serialization, the numerical place and transition identifiers are randomly permuted, preventing the order in which the net was generated, or any other systematic association between identifiers and the queried property, from leaking into the prompt and introducing spurious correlations.

We additionally evaluate alternative serializations in Appendix~\ref{app:serialization} and observe only modest changes in average accuracy, with a maximum gain of $6.4\%$, while preserving the benchmark's difficulty progression across formats.
Given this robustness, we retain the edge list as the default representation because its compact syntax substantially reduces prompt length.

\subsection{Evaluation Protocol}\label{app:evaluation_protocol}
\subsubsection{Model Inference}
Each benchmark question is evaluated independently in a zero-shot setting.
The model receives a system prompt defining all required Petri net definitions and firing semantics, specified in Appendix~\ref{app:system_prompts}, followed by a user prompt containing the task instruction and the serialized net.
No solved examples, conversation history, external tools, or feedback from other questions are provided.
The main experiments use the chain-of-thought templates in Appendix~\ref{app:task_prompts}, which ask the model to provide its reasoning followed by an explicitly marked final answer.
The corresponding no-chain-of-thought evaluation is reported in Appendix~\ref{app:prompting}.

We obtain a single completion per question without self-consistency, majority voting, or repeated sampling, retrying only transient API failures until a successful completion is returned.

Table~\ref{tab:inference-configuration} reports the sampling temperature and maximum output length used for each model configuration.
We retain provider-default sampling parameters as equivalent low-temperature settings are not available across all providers, avoiding asymmetric sampling configurations that could bias cross-model comparisons.
Outputs are capped at $128{,}000$ tokens or the maximum supported by the provider, whichever is lower.
For Claude Haiku~4.5, the extended configuration allocates up to $60{,}000$ tokens to thinking within a $64{,}000$-token output ceiling.

\begin{table*}[t]
    \centering
    \small
    \setlength{\tabcolsep}{6pt}
    \begin{tabular}{lcc}
        \toprule
        Model & Temperature & Maximum Output Tokens \\
        \midrule
        Claude Opus 5       & $1.0$   & $128{,}000$ \\
        Claude Sonnet 5     & $1.0$   & $128{,}000$ \\
        Claude Haiku 4.5    & $1.0$   & $64{,}000$  \\
        GPT-5.6 Sol         & $1.0$   & $128{,}000$ \\
        GPT-5.6 Terra       & $1.0$   & $128{,}000$ \\
        GPT-5.6 Luna        & $1.0$   & $128{,}000$ \\
        Grok 4.6            & $0.7$   & $128{,}000$ \\
        Grok 4.5            & $0.7$   & $128{,}000$ \\
        Grok 4.3            & $0.7$   & $128{,}000$ \\
        Gemini 3.8 Flash    & $1.0$ & $65{,}536$  \\
        Gemini 3.1 Pro      & $1.0$ & $65{,}536$  \\
        Gemma 4 31B         & $1.0$   & $128{,}000$ \\
        GPT-OSS-120B        & $1.0$   & $16{,}000$  \\
        DeepSeek V4 Flash   & $1.0$   & $128{,}000$ \\
        GLM-5               & $1.0$   & $128{,}000$ \\
        GLM-5.3 Flash       & $1.0$   & $128{,}000$ \\
        Mistral Large 3     & $0.7$   & $32{,}000$  \\
        Nemotron 3 Super    & $1.0$   & $32{,}000$  \\
        Kimi K2.5           & $1.0$   & $16{,}000$  \\
        DeepSeek V3.2       & $1.0$   & $8{,}192$   \\
        Qwen3-235B-A22B     & $0.7$   & $8{,}192$   \\
        \bottomrule
    \end{tabular}
    \caption{
        Inference sampling configurations used for the main evaluation in Section~\ref{ssec:benchmark_performance}.
        Maximum output tokens is set to the minimum of $128{,}000$ tokens and the maximum output length supported by the corresponding provider.
        Temperatures correspond to the documented default of each model or endpoint.
    }
    \label{tab:inference-configuration}
\end{table*}

\subsubsection{Parsing and Scoring}\label{app:parsing}

The PetriBench chain-of-thought prompts require a final answer line containing either a Boolean value or an integer, depending on the question type.
Predictions are extracted automatically using a deterministic parser.
For Boolean questions, the parser searches case-insensitively for the phrase ``Final Answer:'' followed by either \emph{True} or \emph{False}; for integer-valued questions, it searches for the same marker followed by an integer.
Whitespace surrounding the marker and answer value is ignored.

To accommodate responses that provide a valid answer but omit the requested marker, we apply a single fallback rule in which the parser extracts the final standalone Boolean value for Boolean questions or the final standalone integer for integer-valued questions.
The same rule supports the direct-answer templates used in the no-chain-of-thought ablation.
Beyond case-insensitive Boolean matching and integer-string conversion, no semantic normalization or model-based interpretation is applied.

A prediction is scored as correct only when the parsed Boolean or integer
exactly matches the ground-truth answer.
The preceding reasoning is not graded, and no partial credit is awarded meaning that responses from which no valid answer can be extracted are recorded as parse failures and counted as incorrect rather than excluded from the evaluation denominator.
Likewise, truncated responses receive credit only when the available text contains a parseable and correct answer.
We release all raw outputs, parsed predictions, and parse-failure indicators to make the scoring procedure fully auditable.

\subsection{Prompts}\label{app:prompts}

\definecolor{pastelcoral}{HTML}{E6A4A4}
\definecolor{pastelblue}{HTML}{A7C7E7}
\definecolor{pastelgreen}{HTML}{A8D5BA}

\newtcolorbox{systemprompt}{
    enhanced,
    breakable,
    colback=white,
    colframe=white,
    boxrule=0pt,
    borderline west={2pt}{0pt}{pastelcoral!85!black},
    left=10pt,
    right=0pt,
    top=2pt,
    bottom=2pt,
    before upper={
        \textbf{\color{pastelcoral!65!black}System Prompt}
        \par\medskip
    },
    before skip=0.7em,
    after skip=0.9em
}

\newtcolorbox{cotprompt}[1]{
    enhanced,
    breakable,
    colback=white,
    colframe=white,
    boxrule=0pt,
    borderline west={2pt}{0pt}{pastelblue!85!black},
    left=10pt,
    right=0pt,
    top=2pt,
    bottom=2pt,
    before upper={
        \textbf{\color{pastelblue!65!black}#1}
        \par\medskip
    },
    before skip=0.7em,
    after skip=0.9em
}

\newtcolorbox{nocotprompt}[1]{
    enhanced,
    breakable,
    colback=white,
    colframe=white,
    boxrule=0pt,
    borderline west={2pt}{0pt}{pastelgreen!85!black},
    left=10pt,
    right=0pt,
    top=2pt,
    bottom=2pt,
    before upper={
        \textbf{\color{pastelgreen!65!black}#1}
        \par\medskip
    },
    before skip=0.7em,
    after skip=0.9em
}

\subsubsection{System Prompt}\label{app:system_prompts}

\begin{systemprompt}
You are an expert Petri net analyzer. You are given a Petri net and a question.

A Petri net has places that hold tokens, transitions that may fire, and arcs that define how many tokens are consumed or produced.

An arc going from a place to a transition consumes a number of tokens equal to the multiplicity. An arc going from a transition to a place produces a number of tokens equal to the multiplicity.

A marking is simply the current distribution of tokens across all places.

A transition can fire only if all its input places contain at least the required number of tokens, simultaneously consuming the required tokens from all inputs and producing tokens at all outputs.

A transition is L0 (dead) if it can never fire in any valid sequence from the initial marking.

A transition is L4 (live) if, from every reachable marking, there exists a valid sequence that will eventually fire it.

A Petri net deadlocks if there is at least one reachable marking where no transitions can fire.

A Petri net is unbounded if there exists a valid sequence of firings that allows the number of tokens in at least one place to grow infinitely large.
\medskip
\medskip

The Petri net will be given as plain text. Each line defines one object.

\medskip
\medskip
place \textless id\textgreater{} \textless tokens\textgreater{}\\
trans \textless id\textgreater{}\\
arc \textless source\textgreater{} \textless target\textgreater{}
\textless multiplicity\textgreater{}

\medskip
\medskip
Your job is to read the Petri net and complete the task.
\end{systemprompt}

\subsubsection{Task Prompts}\label{app:task_prompts}

\begin{cotprompt}{Minimum Token Steps (CoT)}
Determine the minimum number of transition firings needed for place
\texttt{\textless TARGET\textgreater{}} to contain at least
\texttt{\textless THRESHOLD\textgreater{}} tokens.

\medskip

1. Start from the initial marking.

2. Reason only about valid transition firings.

3. Find the earliest reachable marking where place
\texttt{\textless TARGET\textgreater{}} contains at least
\texttt{\textless THRESHOLD\textgreater{}} tokens.

4. Conclude your answer with a single integer.

\medskip
\medskip

**Output Format:**

Reasoning: \textless explanation of thoughts\textgreater{}

Final Answer: [integer]
\end{cotprompt}

\begin{cotprompt}{Reachable Markings (CoT)}
Within \texttt{\textless HORIZON\textgreater{}} transition firings, determine how many distinct markings of the Petri net are reachable from the initial marking, including the initial marking.

\medskip

1. Reason about the markings reachable from the initial marking in at most
\texttt{\textless HORIZON\textgreater{}} firings.

2. Count each distinct marking only once, even if it can be reached by multiple firing sequences.

3. Include the initial marking as reachable.

4. Conclude your answer with the number of distinct reachable markings.

\medskip
\medskip

**Output Format:**

Reasoning: \textless explanation of thoughts\textgreater{}

Final Answer: [integer]
\end{cotprompt}

\begin{cotprompt}{$L_0$ Liveness (CoT)}
Determine if transition \texttt{\textless TRANSITION\textgreater{}} is $L_0$ (dead).

\medskip

1. Analyze the requirements to fire \texttt{\textless TRANSITION\textgreater{}}.

2. Reason about how tokens can be produced, consumed, and cycled in relation to this transition.

3. Determine if there is any valid firing sequence that can ever fire
\texttt{\textless TRANSITION\textgreater{}}. If not, it is $L_0$.

4. Conclude your answer with 'True' or 'False'.

\medskip
\medskip

**Output Format:**

Reasoning: \textless explanation of thoughts\textgreater{}

Final Answer: [True/False]
\end{cotprompt}

\begin{cotprompt}{$L_4$ Liveness (CoT)}
Determine if transition \texttt{\textless TRANSITION\textgreater{}} is $L_4$ (live).

\medskip

1. Analyze the requirements to fire \texttt{\textless TRANSITION\textgreater{}}.

2. Reason about how tokens can be produced, consumed, and cycled in relation to this transition.

3. Determine if, from every possible terminal set of markings, there remains a sequence that can fire
\texttt{\textless TRANSITION\textgreater{}}. If yes, it is $L_4$.

4. Conclude your answer with 'True' or 'False'.

\medskip
\medskip

**Output Format:**

Reasoning: \textless explanation of thoughts\textgreater{}

Final Answer: [True/False]
\end{cotprompt}

\begin{cotprompt}{Deadlock (CoT)}
Determine if the Petri net can reach a deadlock state.

\medskip

1. Analyze the initial marking and possible firing sequences.

2. Identify if there is any sequence of firings that leads to a marking where no transitions have sufficient input tokens to fire.

3. Conclude your answer with 'True' or 'False'.

\medskip
\medskip

**Output Format:**

Reasoning: \textless explanation of thoughts\textgreater{}

Final Answer: [True/False]
\end{cotprompt}

\begin{cotprompt}{Boundedness (CoT)}
Determine if the Petri net is bounded.

\medskip

1. Analyze token production and consumption across all transitions.

2. Look for loops or firing sequences that strictly increase the total number of tokens in a place without a bound.

3. If a marking M' is reachable from M, and M' strictly dominates M (M' > M), the net is unbounded (False). Otherwise, it is bounded (True).

4. Conclude your answer with 'True' or 'False'.

\medskip
\medskip

**Output Format:**

Reasoning: \textless explanation of thoughts\textgreater{}

Final Answer: [True/False]
\end{cotprompt}

\begin{nocotprompt}{Minimum Token Steps (No-CoT)}
What is the minimum number of firings needed for place
\texttt{\textless TARGET\textgreater{}} to contain at least
\texttt{\textless THRESHOLD\textgreater{}} tokens? Answer only with an integer.
\end{nocotprompt}

\begin{nocotprompt}{Reachable Markings (No-CoT)}
Within \texttt{\textless HORIZON\textgreater{}} transition firings, how many distinct markings of the Petri net are reachable from the initial marking, including the initial marking? Answer only with an integer.
\end{nocotprompt}

\begin{nocotprompt}{$L_0$ Liveness (No-CoT)}
Is transition \texttt{\textless TRANSITION\textgreater{}} $L_0$ (dead)? Answer only with 'True' or 'False'.
\end{nocotprompt}

\begin{nocotprompt}{$L_4$ Liveness (No-CoT)}
Is transition \texttt{\textless TRANSITION\textgreater{}} $L_4$ (live)? Answer only with 'True' or 'False'.
\end{nocotprompt}

\begin{nocotprompt}{Deadlock (No-CoT)}
Does the system ever reach a deadlock state (a marking where no transitions are fireable)? Answer only with 'True' or 'False'.
\end{nocotprompt}

\begin{nocotprompt}{Boundedness (No-CoT)}
Is the Petri net bounded (is the token count in every place strictly finite for all reachable markings)? Answer only with 'True' or 'False'.
\end{nocotprompt}

\end{document}

%% file: math_commands.tex
\usepackage{amsmath,amsfonts,bm}

\def\eqref#1{equation~\ref{#1}}

\def\1{\bm{1}}

\DeclareMathAlphabet{\mathsfit}{\encodingdefault}{\sfdefault}{m}{sl}
\SetMathAlphabet{\mathsfit}{bold}{\encodingdefault}{\sfdefault}{bx}{n}

